\pdfoutput=1

\documentclass[11pt]{article}

\usepackage{acl}

\usepackage{times}
\usepackage{latexsym}

\usepackage{algorithm}
\usepackage{algpseudocode}
\usepackage{graphicx}
\usepackage{booktabs}
\usepackage{float}
\usepackage{multicol}
\usepackage{multirow}
\usepackage{graphicx}
\usepackage{amssymb}
\usepackage{xcolor}
\usepackage{pifont}
\usepackage{enumitem}
\usepackage{dblfloatfix}

\newcommand{\cmark}{\ding{51}}%
\newcommand{\xmark}{\ding{55}}%
\definecolor{pos}{RGB}{212,17,89}
\definecolor{neg}{RGB}{53,114,239}
\definecolor{red}{RGB}{212,17,89}
\definecolor{green}{RGB}{91,179,24}
\definecolor{lightgreen}{RGB}{190,231,166}
\definecolor{coral}{RGB}{253,124,91}
\definecolor{navy}{RGB}{86,111,157}
\definecolor{amber}{RGB}{252,200,0}
\definecolor{blue}{RGB}{53,114,239}
\definecolor{grey}{RGB}{199,200,204}

\newcommand{\tool}{\textsc{GEDI}}
\newcommand{\surveyed}{52}

\usepackage[T1]{fontenc}

\usepackage[utf8]{inputenc}

\usepackage{microtype}

\usepackage{inconsolata}

\title{An Electoral Approach to Diversify LLM-based \\Multi-Agent Collective Decision-Making}

\author{Xiutian Zhao \\ University of Edinburgh \\ \texttt{x.zhao-103@sms.ed.ac.uk}
        \And  Ke Wang, Wei Peng  \\ Huawei IT Innovation and Research Center \\ \texttt{\{wangke215, peng.wei1\}@huawei.com}}

\begin{document}
\maketitle

\begin{abstract}
Modern large language models (LLMs) have exhibited cooperative synergy on complex task-solving, and collective decision-making (CDM) is a pivotal component in LLM-based multi-agent collaboration frameworks.
Our survey on \surveyed{} recent such systems uncovers a severe lack of diversity, with a heavy reliance on \textit{dictatorial} and \textit{plurality voting} for CDM.
Through the lens of social choice theory, we scrutinize widely-adopted CDM methods and identify their limitations.
To enrich current landscape of LLM-based CDM, we present \tool{}, an electoral CDM module that incorporates various ordinal preferential voting mechanisms.
Our empirical case study across three benchmarks shows that the integration of certain CDM methods can markedly improve the reasoning capabilities and robustness of some leading LLMs, all without requiring intricate system designs.
Additionally, we find that some CDM mechanisms generate positive synergies even with as few as three agents. The voting-based methods also demonstrate robustness against single points of failure, as well as diversity in terms of hit-rate@k and subject-wise impacts.\footnote{Our code and data are available at \url{https://github.com/xiutian/GEDI}}
\end{abstract}

\section{Introduction}
While multi-agent systems have constantly garnered attention even before the advent of large language models (LLMs), as evidenced by prior works \cite{wooldridge2009introduction, dorri2018multi}. The recent advancements in LLMs have significantly sparked interest in LLM-based agents.
Furthermore, novel techniques such as effective prompt engineering \cite{wei2023chainofthought, wang2023selfconsistency} and agent-interaction schemes \cite{yao2023react, shinn2023reflexion} have propelled a surge in research on collaborative LLM agents \cite{xi2023rise, wang2023survey}.
Researchers have deployed LLM-based agents in various environments and scenarios: from simulating small community \cite{liu2023training, park2023generative} to predicting court judgement \cite{hamilton2023blind}, crafting digital avatars \cite{jarrett2023language, yang2024llm}, and participating in dialogue-based games \cite{xu2023exploring, stepputtis-etal-2023-long, li-etal-2023-theory}, among others.

\begin{figure}[t]
  \begin{minipage}[c]{0.42\columnwidth}
    \caption{\label{fig:mas_distribution}
       Distribution of CDM methods in \surveyed{} LLM-based multi-agent collaboration systems, denoting a severe lack of diversity.
    }
  \end{minipage}\hfill
  \begin{minipage}[c]{0.56\columnwidth}
    \includegraphics[width=\columnwidth]{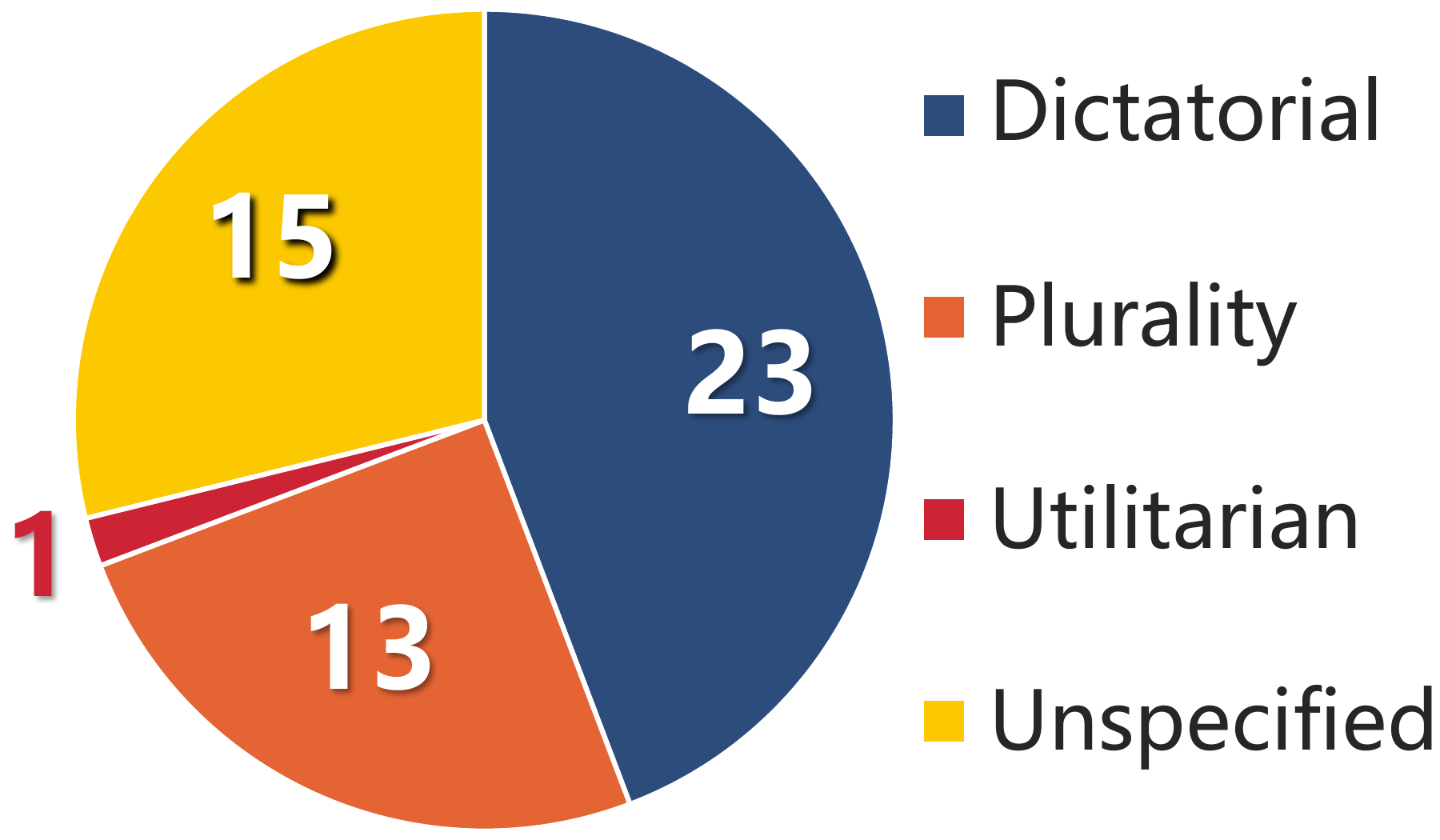}
  \end{minipage}
\end{figure}

However, the existing accounts on LLM-based multi-agent collaboration has been heavily focusing on inter-agent communication and interaction workflows. In contrast, another vital aspect, collective decision-making (CDM), appears to have been largely neglected and overly simplified. 
Our review of \surveyed{} recent LLM collaboration systems (\S~\ref{sec:survey}) reveals that systems either appoint a `dictator' agent to make decisions for the group \cite{hao2023chatllm, nair2023dera} or depend on simplistic plurality voting \cite{chan2023chateval, zhang2023exploring, xu2023reasoning}, with one case adopting an utilitarian approach \cite{jarrett2023language}.

This study examines prevalent CDM methods through the lens of social choice theory \cite{arrow2010handbook} and illustrate their failure to meet fundamental criteria (\S~\ref{sec:SCT}): \textit{dictatorial} methods are fragile for their absolute dependency on one single agent; \textit{plurality voting}, while simple and intuitively flawless, disqualifies \textit{Independence from Irrelevant Alternatives (IIA)} and \textit{Condorcet} criterion; \textit{utilitarian} violates both \textit{Majority} and \textit{Condorcet} criteria.
Such deviations from key criteria may impede the transition from individual preferences to collective decisions among LLM-based agents.

While Arrow's theorems \cite{arrow1951social} establishes the axiomatic impossibility of designing a perfect voting-based CDM system, we can still circumvent some limitations and risks by incorporating a variety of CDM methods into LLM-based multi-agent frameworks. 
To this end, we develop an electoral CDM module, \tool{} (\S~\ref{sec:gedi}), which offers a range of CDM mechanisms that were not previously tested in such frameworks.
To evaluate the potential impact of various CDM methods, we conduct an empirical case study (\S~\ref{sec:case study}) on three multiple-choice question-answering (MCQA) benchmarks: MMLU \cite{hendrycks2021measuring}, MMLU-Pro \cite{wang2024mmlupro}, and ARC-Challenge \cite{clark2018think},  using a suite of models with various sizes and architectures.

Our key findings (\S~\ref{sec:results}) are as follows: 
(1) applying a CDM method generally leads to better results compared to a single-agent decision-making on MCQA benchmarks, though at the cost of increased computation; 
(2) the degree of synergy depends significantly on the backbone model and the benchmark. Some LLMs exhibit substantial improvements with voting-based methods, while others show little to no effect under any CDM; 
(3) most voting methods require only a minimal quorum, as few as three agents, to be effective; 
and (4) CDM methods exhibit varying levels of robustness against unreliable agents, different hit-rates@$k$, and varying impacts across different subject domains.

We hope these observations will encourage further evaluation of the effectiveness of LLM-based multi-agent frameworks and provide valuable insights for advancing LLM-based Multi-Agent Systems (MAS).

\section{A Concise Survey on LLM-based Multi-Agent Collective Decision-Making}\label{sec:survey}
\subsection{Background}
Multi-agent systems are composed of multiple computing elements with autonomous action and interaction capabilities (i.e., `agent') \cite{wooldridge2009introduction}. Prior to the advent of LLMs, research on multi-agent systems had already been a focal point various across disciplines  \cite{silver2017mastering, dorri2018multi}. The swift progression of LLMs has since ignited an intensified interest in employing LLMs as agents \cite{xi2023rise}.
Notably, the advent of effective prompting schemes has greatly boosted the performance of individual LLM agent: Chain-of-Thought \cite{wei2023chainofthought}, Self-Consistency \cite{wang2023selfconsistency}, ReAct \cite{yao2023react}, Reflexion \cite{shinn2023reflexion}, DiVeRSe \cite{li-etal-2023-making} among others. 
Although single-agent frameworks have shown remarkable success in certain NLP tasks, they often struggle with more intricate challenges, such as common sense reasoning and long-term planning \cite{wang2023survey}. In response, some researchers advocate multi-LLM-agent collaboration as a promising path.

\subsection{Collective Decision-Making in LLM-based Multi-Agent Collaboration}
Collective decision-making (CDM) is the process by which a group of autonomous entities arrives at a decision \cite{bose2017collective}. This phenomenon is prevalent in both animal societies and human communities, with numerous interdisciplinary studies corroborating that CDM typically yields superior decisions compared to those made by individuals alone \cite{king2007use, couzin2011uninformed}.

Recent development of LLMs has made self-governing CDM processes feasible in LLM-based multi-agent systems. However, our survey of \surveyed{} newly proposed frameworks indicates that CDM mechanisms have not received adequate focus. Specifically, most systems either depend on the \textit{dictatorial} judgment of a single agent (often by preassigned role) or employ \textit{plurality voting} for decision-making. As depicted in Figure~\ref{fig:mas_distribution}, we can categorize current LLM-based multi-agent systems into four groups based on their CDM approaches: (1) \textit{dictatorial}, (2) \textit{plurality}, (3) \textit{utilitarian}, and (4) those with no CDM or unspecified.

\paragraph{Dictatorial}
Among the reviewed papers, \textit{dictatorial} methods are most popular. As the name implies, it is a one-agent-rule system in which a single agent, often pre-designated, has the right to ratify a decision. Nonetheless, such system can be `collective' in a sense that the `dictator' may be counseled by and communicated with other agents.

Most \textit{dictatorial} frameworks designate a special agent who oversees collaboration, evaluates outcomes, and has the final say over system-level decisions. Such agent has many alias: `leader' \cite{hao2023chatllm, darcy2024marg}, `decider' \cite{nair2023dera}, `commander' \cite{wu2023autogen}, `critic' \cite{li2023camel}, `teacher' \cite{jinxin2023cgmi}, `judge' \cite{liang2023encouraging, xiong-etal-2023-examining, sun2023corex, talebirad2023multiagent}, `evaluator' \cite{tang2023causalgpt}, `planner' \cite{zhang2023proagent, fang2024multiagent}, `recruiter' \cite{li2023metaagents}, `inspector'\cite{hua2024trustagent, wang2024xuatcopilot}, `discriminator' \cite{hang2024cca}, `task agent'\cite{li2023traineragent}, `QA-Checker' \cite{tang2024collaborative}.
Some specific cases include creating virtual software and game development companies hosting LLM-agents of various roles to achieve rapid and low-cost development of software \cite{qian2023communicative, chen2023gamegpt}.
Specially, \citet{chen2023autoagents} suggest an `oligarchic' small group of `planner' and `observers' instead of a single decision-maker.

\paragraph{Plurality Voting}
\textit{Plurality voting} selects the option with the most first-preference votes (i.e., relative majority).
For simplicity, we consider \textit{majority voting}, which that requires more-than-half votes (i.e., absolute majority), and \textit{consensus}, which demands an unanimous agreement from every agents, to be two variations of \textit{plurality voting}.

Frameworks that adapt \textit{plurality voting} often introduce multi-round discussion to reach resolution or majority agreement \cite{xu2023reasoning}.
Multi-agent debate process is found to improve LLMs' factuality \cite{du2023improving}, reasoning capabilities \cite{zhang2023exploring}, and financial trading performances \cite{li2023tradinggpt}.
\citet{chan2023chateval} also improve the quality of evaluation provided by LLM-agents on natural language generation tasks via debates.  
\citet{chen2023agentverse} fashion automatic team formation and LLM-agent experts recruitment.
\citet{chen2023multiagent} quantify consensus-seeking process by appending self-assigned `state' values of LLM-agents and measuring their convergence.
Notably, \citet{wang2023unleashing} showcase that multiple `personas' of a single LLM can also `self-collaborate'.
In some cases, \textit{plurality voting} is chosen to match simulated target scenarios.
\citet{hamilton2023blind} trains nine separated agents as judges to simulate the U.S. Supreme Court and achieve better-than-random judgement prediction accuracy on 96 real-world cases.
In textual or conceptual games like Werewolf \cite{xu2023exploring} and Avalon \cite{stepputtis-etal-2023-long, shi2023cooperation}, agents are bound by the game rule to take this method.

\paragraph{Utilitarian}
\textit{Utilitarian} approaches quantify the impacts of possible decisions and choose the one that maximizes the collective `utility' or `reward' gained by a group. However, \textit{utilitarian} is distinct from other methods for its non-self-governing: the utilities are externally predetermined or updated.
\citet{jarrett2023language} propose to train LLM agents as digital proxy to represent individual preferences via an utilitarian `payoff function'.
Although \textit{utilitarian} is rare in newly proposed LLM-based frameworks, it is a pillar method in many previous non-LLM multi-agent systems \cite{dorri2018multi}.

\paragraph{No CDM or Unspecified}
Some multi-agent scenarios necessitate no CDM. For instance, simulating social interaction and behaviors among LLM-agents \cite{park2023generative, liu2023training, ghaffarzadegan2023generative, hua2023war, zhang2024speechagents, wei2024editable}, while one-to-one agreement can happen occasionally. 
Other systems intrinsically deny a CDM process, such as strictly linear collaboration workflow \cite{hong2023metagpt, wang2023macsql, ding2023designgpt, rasheed2024codepori} or decentralized team arrangements \cite{li-etal-2023-theory, nakajima2023babyagi, he-etal-2023-lego}.
In addition, some frameworks involve human judgement for system-level decisions \cite{ghafarollahi2024protagents, NI2024102131}.

Thus far, having seen a great lack of diversity of CDM methods in LLM-based multi-agent collaboration, we draw our inspiration from social choice theory and scrutinize the pros and cons of the widely-used methods.

\section{A Social Choice Theory Perspective on Collective Decision-Making}\label{sec:SCT}
Social choice theory concerns passing from individual preferences to collective decisions \cite{arrow2010handbook}. While humans have practiced and refined collective decision-making since antiquity, modern social choice theory has not been established until the publishing of Kenneth J. Arrow's renowned \textit{Social Choice and Individual Values} \cite{arrow1951social}, which axiomatically formalizes the theory and comparatively analyzes various electoral systems. 

\subsection{Related Work Incorporating Social Choice Theory into NLP Research}
The related research to date has tended to focus on integrating social choice theory into model alignment \cite{mishra2023ai}, model ensemble \cite{jiang-etal-2023-llm}, text generation and preference extrapolation \cite{fish2023generative}. More specifically, \citet{jarrett2023language} take an \textit{utilitarian} approach to employ LLM agents as digital representatives of human.
\citet{irurozki2022best, rofin-etal-2023-votenrank} point out the limitations of canonical mean-average aggregation of multi-task scores in NLP benchmarking and propose novel aggregation methods based on social choice theory.
\citet{wang2023selfconsistency, xue-etal-2023-dynamic} propose to select answers from multiple generated reasoning paths by \textit{plurality voting} and yield improved results over \textit{utilitarian} approaches.

Most recently, \citet{li2024agents} demonstrate the synergy of \textit{plurality voting} on \texttt{gpt-3.5} \cite{ouyang2022training} and Llama-2 \cite{touvron2023llama}, echoing some of our findings, yet it lacks comparisons with other CDM methods. Another concurrent work \cite{yang2024llm} examines the differences between human and LLM from a voting behavior perspective.
Nevertheless, previous studies do not overlap with our primary aim of diversifying LLM-based multi-agent CDM methods.

\subsection{Criticism on Prevalent CDM Methods in LLM-based Multi-Agent Collaboration}\label{sec:criticism}
In the context of LLM-agent collaboration, \textit{dictatorial} methods rely on a single agent who is informed and counseled by other agents to decide for the group. While dictatorship is often computing-wise efficient, its absolute dependency on a sole agent makes it is more biased and less robust than more `democratic' processes.

In contrast, \textit{utilitarian} and \textit{cardinal voting} methods certainly aggregate and disclose broader individual preferences from group members.
However, an unignorable drawback of these methods is the unstable and arbitrary nature of externally imposed utilities \cite{brandt2016handbook}. Provided that agents have accurate cardinal utilities over choices, which is a strong assumption, then an uneven distribution of utilities is another potential concern: such a system could easily violate \textit{Majority} criterion (see Figure~\ref{fig:utilitarian_majority}) or even collapse to autocratic if one agent with dominant utility impact was present.

\textit{Plurality voting} showcases a paradigmatic example of \textit{ordinal voting} (also known as preferential or ranked voting), another decentralized decision-making family.
Although there are other widely-practiced ordinal voting methods available, to the best of our knowledge, all existing LLM-agent collaboration frameworks that employ voting methods select \textit{plurality voting}, as shown in Figure~\ref{fig:mas_distribution}.
The simple method may seem intuitively `safe'. However, through the lens of Arrow's theorems \cite{arrow1951social}, this method contradicts some rather self-evident criteria. To name two, the method violates both \textit{Independence from Irrelevant Alternatives (IIA)} criterion, as shown in Figure~\ref{fig:plurality_IIA}, and \textit{Condorcet} criterion, as illustrated in Figure~\ref{fig:plurality_condorcet}.

\begin{table}[t]
\resizebox{\columnwidth}{!}{%
\begin{tabular}{@{}lcccccc@{}}
\toprule
\textbf{\begin{tabular}[c]{@{}l@{}}CDM\\ Method\end{tabular}} &
  \textbf{\begin{tabular}[c]{@{}c@{}}Major\\ -ity\end{tabular}} &
  \textbf{\begin{tabular}[c]{@{}c@{}}Mono\\ -tonic\end{tabular}} &
  \textbf{\begin{tabular}[c]{@{}c@{}}Consis\\ -tency\end{tabular}} &
  \textbf{IIA} &
  \textbf{\begin{tabular}[c]{@{}c@{}}Cond\\ -orcet\end{tabular}} &
  \textbf{\begin{tabular}[c]{@{}c@{}}Ballot\\ type\end{tabular}} \\ \midrule
Dictatorial (Blind) & \color{red}\xmark   & \color{green}\cmark & \color{green}\cmark & \color{green}\cmark & \color{red}\xmark   & Ranking \\
Range Voting        & \color{red}\xmark   & \color{green}\cmark & \color{green}\cmark & \color{green}\cmark & \color{red}\xmark   & Scores  \\
Plurality           & \color{green}\cmark & \color{green}\cmark & \color{green}\cmark & \color{red}\xmark   & \color{red}\xmark   & Single* \\
Borda Count         & \color{red}\xmark   & \color{green}\cmark & \color{green}\cmark & \color{red}\xmark   & \color{red}\xmark   & Ranking \\
IRV                 & \color{green}\cmark & \color{red}\xmark   & \color{red}\xmark   & \color{red}\xmark   & \color{red}\xmark   & Ranking \\
Ranked Pairs        & \color{green}\cmark & \color{green}\cmark & \color{red}\xmark   & \color{red}\xmark   & \color{green}\cmark & Ranking \\ \bottomrule
\end{tabular}%
}
\caption{Criteria compliance of some typical CDM methods. \textit{Range Voting} can be viewed as a special \textit{utilitarian} method. \textbf{IIA} denotes \textit{Independence from Irrelevant Alternatives}. *Single ballots can be derived from ranking ones. Find some examples in Appendix~\ref{appendix:criteria_example}.
See Figure~\ref{fig:irv_monotonic} for an example of \textit{instant-runoff voting (IRV)} disqualifying \textit{monotonic} criterion.
}
\label{table:vote_crit}
\end{table}

In fact, Arrow's theorems mathematically prove that every electoral system has some fundamental flaws, as exemplified in Table~\ref{table:vote_crit}. The axiomatic impossibility of constructing a perfect voting system, however, motivates us to reduce the risk of falling into a single point of failure.
To this end, we argue it is of great pragmatic values to diversify current landscape of LLM-agent with modern decentralized voting systems.
In order to leverage LLM-agents' natural-language-based `judgement' rather than imposed `utility' or `reward', we place a particular emphasis on ordinal preferential voting.

\section{Diversifying LLM-based Multi-Agent CDM}\label{sec:gedi}
\begin{figure*}[h]
	\begin{center}
		\includegraphics[width=\linewidth]{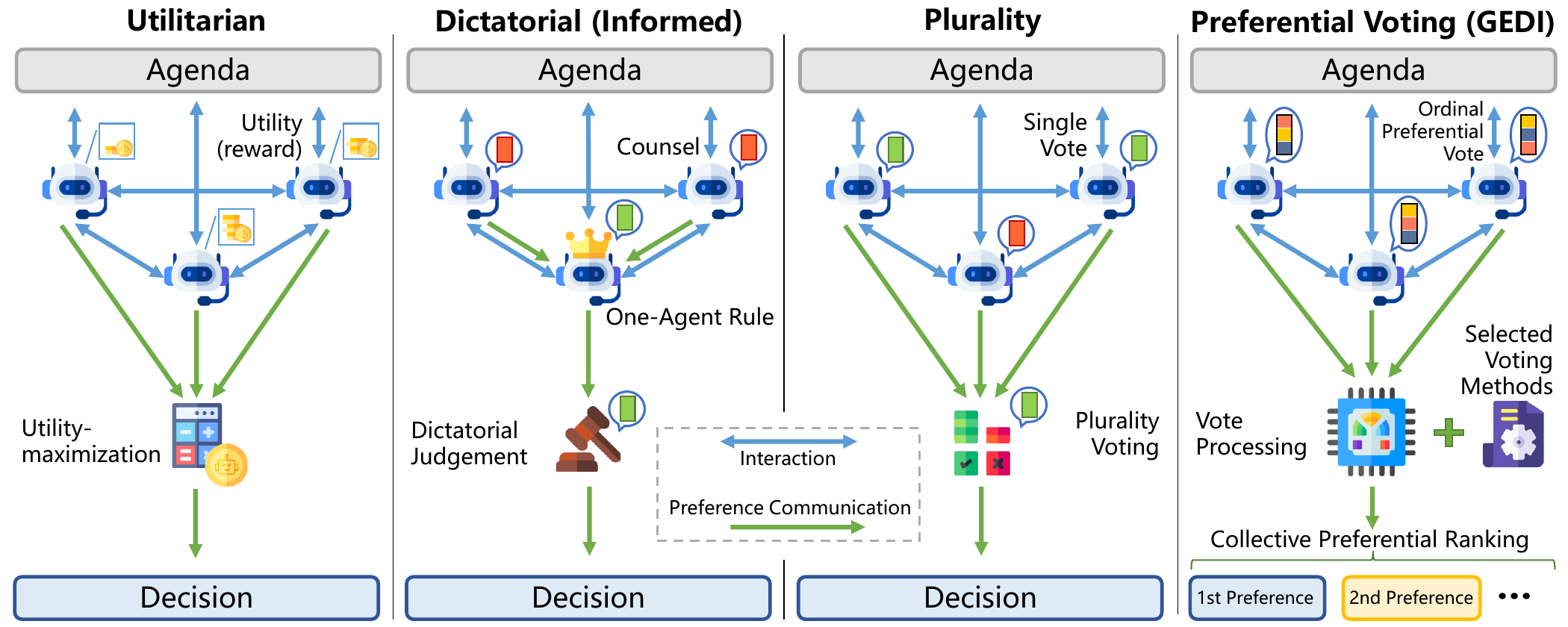}
	\end{center}
	\caption{
    Comparison among different LLM-based multi-agent CDM structures: utilitarian, dictatorial, plurality and our expansion. Agenda refers to assigned tasks or interactive environment. \textcolor{blue}{Blue} and \textcolor{green}{green} arrows denote interaction between agents and preference communication to CDM systems respectively. Rather than generate a single decision, \tool{} uniquely outputs ordinal rankings, providing more information on agents' collective preferences.
    }
	\label{fig:gedi}
\end{figure*}

To enhance the diversity of CDM approaches within LLM-agent frameworks, we propose incorporating a range of CDM methods rooted in human socio-political practices.
Specifically, we craft an electoral CDM module, named \textbf{G}eneral \textbf{E}lectoral \textbf{D}ecision-making \textbf{I}nterface (\tool{}), which integrates several common ordinal preferential voting systems. Figure~\ref{fig:gedi} highlights a few key distinctions between \tool{} and other commonly used CDM methods in LLM-based MAS.

\subsection{Definition}\label{sec:definiation}
Following conventional practice \cite{arrow2010handbook, brandt2016handbook}, consider a multi-alternative decision-making process. Let $N=\{1,2,...,n\}$ be a finite set of $n$ agents,
$A = \{a_1, a_2,...a_m\}$ be a finite set of $m$ distinct alternatives, where $m \geq 2$ and for all $a,b \in A, a \ne b$.
A preferential ranking \textit{ballot} (i.e., vote) can be defined as a \textit{strict partial ordering} $\succ$ of $A$ \cite{rosen2007discrete}. Specifically, $\succ$ is \textit{transitive}: for all $a,b,c \in A$, if $a \succ b$ and $b \succ c$ then $a \succ c$); and \textit{complete}: for all $a,b \in A$, $a \succ b$ or $a \prec b$. Note that there is also a \textit{weak ordering} variation that accepts voters stating indifference to two alternatives (i.e., preferential ties).

Concretely, the input of \tool{} is composed of: (1) a \textit{profile} $P = (\succ_1,\succ_2,...\succ_n)$, which denotes a collection of ballots from each voter $i \in N$; (2) a voting system (i.e., \textit{social choice function (SCF)}), which is defined as a map $f:\mathcal{L}(A)^n \to \mathcal{C}(A)$ that returns a set of alternatives for each profile of strict preferences. The output $f(P)$ is a nonempty ordered subset of the alternative set $A$.

\subsection{Assessed Electoral Methods}\label{sec:methods}
We select 10 CDM methods to assess in the following case study: \textit{Blind Dictatorial}, \textit{Informed Dictatorial}, \textit{Mis-informed Dictatorial}, \textit{Range Voting}, \textit{Plurality}, \textit{Borda Count}, \textit{Bucklin}, \textit{Minimax}, and \textit{Ranked Pairs}, along with random baselines. 

\paragraph{Dictatorial} 
\textit{Blind dictatorial} (or \textit{random ballot}) arbitrarily chooses an agent and admits its preference ranking as the decision without consulting with fellow agents \cite{aziz2013popular}. 
Alternatively, \textbf{informed dictatorial} is a counselled version in which a `dictator' agent first reviews ballots of the voting ensemble and then forms its own decision. 
We also entail a \textbf{mis-informed} variation to verify the impact of information communicated via ballots, in which the `dictator' is consulted by random ballots rather than actual ones from the ensemble.
Notably, \textit{blind dictatorial} still requires a full set of ballots, which distinguishes it from a random baseline that arranges a preference ranking in a stochastic manner without actual voting.
    
\paragraph{Range Voting} Agents rate alternatives under a designated cardinal range, and the winner is selected by highest overall scores \cite{menton2013normalized}. This approach resembles \textit{utilitarian} methods, yet the `utilities' (i.e., overall scores) are given by agents rather than externally assigned. 

\paragraph{Plurality} Simple plurality (relative majority) considers only the first-preference in each vote, ignoring any later preferences. The winner is the candidate who receives the most top-choice votes.

\paragraph{Bucklin Voting}
The first-preference votes are accounted for first, and if no choice has absolute majority, next-in-line preference votes are then accounted. Repeat the process until an absolute majority winner emerge \cite{erdelyi2015control}.

\paragraph{Borda Count} Choices gain points from their places on each ballot, and the overall points determine the winner \cite{emerson2013original, davies2014complexity}. In standard Borda count, a preference ballot on $m$ alternatives awards $m-i$ points for the $i$-th ranked alternative. Unlike range voting, Borda is distinct from \textit{utilitarian} methods because Borda still utilises \textit{ordinal} preferences.
    
\paragraph{Instant-Runoff Voting (IRV)} A multi-round mechanism that repeatedly eliminates the alternative with fewest first-preference votes and `transfers' the votes of the eliminated to surviving alternatives \cite{Freeman_Brill_Conitzer_2014}. The reverse order of elimination comprise the sorted list of collectively preferred options. While there are various early exit designs (e.g., choose a winner once an absolute majority appears), we include the standard  IRV to get the full sorted list.

\paragraph{Minimax} A method that selects the choice with least `worst disfavor' \cite{brams2007minimax}. Formally, let $f(a,b)$ represent the overall `favor' of $a$ over $b$ (i.e., the number of pairwise wins of $a$ against $b$ across all ballots) for a distinct pair of alternatives $a, b \in A$. $f(a,b)$ can be negative if more voters favor $b$ over $a$. The `worst disfavor' of $a$ is defined as $\max f(b,a)$. The winning alternative is the one minimizes the worst disfavor, i.e., the one with the minimum $\max f(b,a)$: \begin{math} a_w = \arg \min \limits_{a} (\max \limits_{b} f(b,a)) \end{math}.

\paragraph{Ranked Pairs}
Concretely, let $(a,b)$ denotes the aggregated pairwise comparison result of $a, b \in A$, and a positive $(a,b)$ indicates more agents prefer $a$ over $b$. Ranked pairs method  breaks down complete ballots into preferential pairs and ranks them by prevalence. Starting from the most frequent pairs (i.e., \begin{math} \arg \max \limits_{a,b \in A} (a,b) \end{math}), the method fills a pairwise comparison matrix, marking the pairs and their transitive results positive, ignoring any conflicting pairs with smaller frequency. The winner is \begin{math} \arg \limits_{a \in A} ((a,b) > 0) \ s.t.\ \forall b \in A, a \neq b \end{math}, the one who has positive signs over all other alternatives.

\begin{table*}[t]
\resizebox{\textwidth}{!}{%
\begin{tabular}{@{}lccccccccccc@{}}
\toprule
\textbf{Base Model} &
  \textbf{Rand.} &
  \textbf{Score} &
  \multicolumn{3}{c}{\textbf{Dictatorial-based}} &
  \multicolumn{6}{c}{\textbf{Ordinal Ranking}} \\ \cmidrule(lr){2-2} \cmidrule(lr){3-3} \cmidrule(lr){4-6} \cmidrule(l){7-12} 
\begin{tabular}[c]{@{}c@{}}\\ \textit{\textbf{MMLU}}\end{tabular} &
  Rand. &
  \begin{tabular}[c]{@{}c@{}}Range\\ Voting\end{tabular} &
  \begin{tabular}[c]{@{}c@{}}Blind\\ Dicta.\end{tabular} &
  \begin{tabular}[c]{@{}c@{}}Informed\\ Dicta.\end{tabular} &
  \begin{tabular}[c]{@{}c@{}}Mis-Informed\\ Dicta.\end{tabular} &
  Plurality &
  Bucklin &
  \begin{tabular}[c]{@{}c@{}}Borda\\ Count\end{tabular} &
  IRV &
  Minimax &
  \begin{tabular}[c]{@{}c@{}}Ranked\\ Pairs\end{tabular} \\ \midrule
\texttt{mistral-7b} &
  24.8 &
  \multicolumn{1}{c|}{51.8 \small{\textcolor{neg}{(-4.6)}}} &
  56.4 &
  55.9 \small{\textcolor{neg}{(-0.5)}} &
  \multicolumn{1}{c|}{36.1 \small{\textcolor{neg}{(-20.3)}}} &
  56.8 \small{\textcolor{pos}{(+0.4)}} &
  57.1 \small{\textcolor{pos}{(+0.7)}} &
  56.9 \small{\textcolor{pos}{(+0.5)}} &
  56.9 \small{\textcolor{pos}{(+0.5)}} &
  57.0 \small{\textcolor{pos}{(+0.6)}} &
  57.0 \small{\textcolor{pos}{(+0.6)}} \\
\texttt{llama-3-8b} &
  25.0 &
  \multicolumn{1}{c|}{37.7 \small{\textcolor{neg}{(-7.3)}}} &
  45.0 &
  36.5 \small{\textcolor{neg}{(-8.5)}} &
  \multicolumn{1}{c|}{32.2 \small{\textcolor{neg}{(-12.8)}}} &
  45.9 \small{\textcolor{pos}{(+0.9)}} &
  46.4 \small{\textcolor{pos}{(+1.4)}} &
  46.3 \small{\textcolor{pos}{(+1.3)}} &
  45.7 \small{\textcolor{pos}{(+0.7)}} &
  45.9 \small{\textcolor{pos}{(+0.9)}} &
  46.0 \small{\textcolor{pos}{(+1.0)}} \\
\texttt{glm-4-9b} &
  25.2 &
  \multicolumn{1}{c|}{61.3 \small{\textcolor{neg}{(-0.4)}}} &
  61.7 &
  54.3 \small{\textcolor{neg}{(-7.4)}} &
  \multicolumn{1}{c|}{\textbf{53.0 \small{\textcolor{neg}{(-8.7)}}}} &
  \textbf{64.6 \small{\textcolor{pos}{(+2.9)}}} &
  \textbf{64.5 \small{\textcolor{pos}{(+2.8)}}} &
  64.1 \small{\textcolor{pos}{(+2.4)}} &
  \textbf{64.9 \small{\textcolor{pos}{(+3.2)}}} &
  \textbf{64.4 \small{\textcolor{pos}{(+2.7)}}} &
  \textbf{64.6 \small{\textcolor{pos}{(+2.9)}}} \\
\texttt{llama-3-70b} &
  25.3 &
  \multicolumn{1}{c|}{74.9 \small{\textcolor{pos}{(+1.6)}}} &
  73.3 &
  70.1 \small{\textcolor{neg}{(-3.2)}} &
  \multicolumn{1}{c|}{62.6 \small{\textcolor{neg}{(-10.7)}}} &
  73.9 \small{\textcolor{pos}{(+0.6)}} &
  73.8 \small{\textcolor{pos}{(+0.5)}} &
  73.7 \small{\textcolor{pos}{(+0.4)}} &
  73.9 \small{\textcolor{pos}{(+0.6)}} &
  73.9 \small{\textcolor{pos}{(+0.6)}} &
  73.9 \small{\textcolor{pos}{(+0.6)}} \\
\texttt{qwen-2-72b} &
  25.1 &
  \multicolumn{1}{c|}{69.2 \small{\textcolor{neg}{(-0.5)}}} &
  69.7 &
  69.7 \small{($\pm$0.0)} &
  \multicolumn{1}{c|}{39.5 \small{\textcolor{neg}{(-30.2)}}} &
  70.0 \small{\textcolor{pos}{(+0.3)}} &
  69.9 \small{\textcolor{pos}{(+0.2)}} &
  70.0 \small{\textcolor{pos}{(+0.3)}} &
  69.9 \small{\textcolor{pos}{(+0.2)}} &
  69.9 \small{\textcolor{pos}{(+0.2)}} &
  69.9 \small{\textcolor{pos}{(+0.2)}} \\
\texttt{qwen-1.5-110b} &
  25.0 &
  \multicolumn{1}{c|}{71.3 \small{\textcolor{neg}{(-1.5)}}} &
  72.8 &
  73.0 \small{\textcolor{pos}{(+0.2)}} &
  \multicolumn{1}{c|}{46.3 \small{\textcolor{neg}{(-26.5)}}} &
  72.9 \small{\textcolor{pos}{(+0.1)}} &
  72.9 \small{\textcolor{pos}{(+0.1)}} &
  72.7 \small{\textcolor{neg}{(-0.1)}} &
  72.9 \small{\textcolor{pos}{(+0.1)}} &
  72.9 \small{\textcolor{pos}{(+0.1)}} &
  72.9 \small{\textcolor{pos}{(+0.1)}} \\
\texttt{gpt-3.5} &
  24.9 &
  \multicolumn{1}{c|}{63.0 \small{\textcolor{pos}{(+2.2)}}} &
  60.8 &
  \textbf{64.7 \small{\textcolor{pos}{(+3.9)}}} &
  \multicolumn{1}{c|}{36.9 \small{\textcolor{neg}{(-23.9)}}} &
  \textbf{65.9 \small{\textcolor{pos}{(+5.1)}}} &
  \textbf{65.5 \small{\textcolor{pos}{(+4.7)}}} &
  \textbf{65.6 \small{\textcolor{pos}{(+4.8)}}} &
  \textbf{65.6 \small{\textcolor{pos}{(+4.8)}}} &
  \textbf{65.6 \small{\textcolor{pos}{(+4.8)}}} &
  \textbf{65.6 \small{\textcolor{pos}{(+4.8)}}} \\
\texttt{gpt-4} &
  25.0 &
  \multicolumn{1}{c|}{\textbf{80.7 \small{\textcolor{pos}{(+5.1)}}}} &
  75.6 &
  \textbf{82.1 \small{\textcolor{pos}{(+6.5)}}} &
  \multicolumn{1}{c|}{\textbf{70.9 \small{\textcolor{neg}{(-4.7)}}}} &
  \textbf{82.5 \small{\textcolor{pos}{(+6.9)}}} &
  \textbf{81.9 \small{\textcolor{pos}{(+6.3)}}} &
  \textbf{81.9 \small{\textcolor{pos}{(+6.3)}}} &
  \textbf{81.9 \small{\textcolor{pos}{(+6.3)}}} &
  \textbf{81.9 \small{\textcolor{pos}{(+6.3)}}} &
  \textbf{81.9 \small{\textcolor{pos}{(+6.3)}}} \\ \midrule
\textit{\textbf{MMLU-Pro}} &
   &
  \multicolumn{1}{l}{\textit{\textbf{}}} &
   &
  \multicolumn{1}{l}{} &
  \multicolumn{1}{l}{} &
   &
   &
   &
   &
   &
   \\ \midrule
\texttt{mistral-7b} &
  9.6 &
  \multicolumn{1}{c|}{20.9 \small{\textcolor{neg}{(-9.0)}}} &
  29.9 &
  27.7 \small{\textcolor{neg}{(-2.2)}} &
  \multicolumn{1}{c|}{15.6 \small{\textcolor{neg}{(-14.3)}}} &
  31.7 \small{\textcolor{pos}{(+1.8)}} &
  30.7 \small{\textcolor{pos}{(+0.8)}} &
  31.4 \small{\textcolor{pos}{(+1.5)}} &
  31.2 \small{\textcolor{pos}{(+1.3)}} &
  31.7 \small{\textcolor{pos}{(+1.8)}} &
  31.7 \small{\textcolor{pos}{(+1.8)}} \\
\texttt{llama-3-8b} &
  9.7 &
  \multicolumn{1}{c|}{18.9 \small{\textcolor{neg}{(-2.4)}}*} &
  21.3 &
  \textbf{23.8 \small{\textcolor{pos}{(+2.5)}}} &
  \multicolumn{1}{c|}{\textbf{19.3 \small{\textcolor{neg}{(-2.0)}}}} &
  22.2 \small{\textcolor{pos}{(+0.9)}} &
  \textbf{23.8 \small{\textcolor{pos}{(+2.5)}}} &
  \textbf{24.5 \small{\textcolor{pos}{(+3.2)}}} &
  22.6 \small{\textcolor{pos}{(+1.3)}} &
  23.0 \small{\textcolor{pos}{(+1.7)}} &
  23.4 \small{\textcolor{pos}{(+2.1)}} \\
\texttt{glm-4-9b} &
  9.6 &
  \multicolumn{1}{c|}{26.2 \small{\textcolor{neg}{(-5.7)}}*} &
  31.9 &
  28.2 \small{\textcolor{neg}{(-3.7)}} &
  \multicolumn{1}{c|}{\textbf{23.9 \small{\textcolor{neg}{(-8.0)}}}} &
  \textbf{36.4 \small{\textcolor{pos}{(+4.5)}}} &
  \textbf{35.9 \small{\textcolor{pos}{(+4.0)}}} &
  \textbf{34.8 \small{\textcolor{pos}{(+2.9)}}} &
  \textbf{36.7 \small{\textcolor{pos}{(+4.8)}}} &
  \textbf{35.6 \small{\textcolor{pos}{(+3.7)}}} &
  \textbf{36.2 \small{\textcolor{pos}{(+4.3)}}} \\
\texttt{llama-3-70b} &
  10.3 &
  \multicolumn{1}{c|}{\textbf{46.7 \small{\textcolor{pos}{(+3.5)}}}} &
  43.2 &
  44.6 \small{\textcolor{pos}{(+1.4)}} &
  \multicolumn{1}{c|}{24.6 \small{\textcolor{neg}{(-18.6)}}} &
  42.8 \small{\textcolor{neg}{(-0.4)}} &
  43.5 \small{\textcolor{pos}{(+0.3)}} &
  43.6 \small{\textcolor{pos}{(+0.4)}} &
  43.0 \small{\textcolor{neg}{(-0.2)}} &
  43.2 \small{($\pm$0.0)} &
  43.5 \small{\textcolor{pos}{(+0.3)}} \\
\texttt{qwen-2-72b} &
  10.4 &
  \multicolumn{1}{c|}{35.1 \small{\textcolor{neg}{(-1.7)}}} &
  36.8 &
  37.4 \small{\textcolor{pos}{(+0.6)}} &
  \multicolumn{1}{c|}{19.5 \small{\textcolor{neg}{(-17.3)}}} &
  37.2 \small{\textcolor{pos}{(+0.4)}} &
  36.7 \small{\textcolor{neg}{(-0.1)}} &
  36.7 \small{\textcolor{neg}{(-0.1)}} &
  37.2 \small{\textcolor{pos}{(+0.4)}} &
  37.3 \small{\textcolor{pos}{(+0.5)}} &
  37.2 \small{\textcolor{pos}{(+0.4)}} \\
\texttt{qwen-1.5-110b} &
  10.1 &
  \multicolumn{1}{c|}{45.7 \small{\textcolor{pos}{(+0.9)}}} &
  44.8 &
  42.8 \small{\textcolor{neg}{(-2.0)}} &
  \multicolumn{1}{c|}{16.6 \small{\textcolor{neg}{(-28.2)}}} &
  44.7 \small{\textcolor{neg}{(-0.4)}} &
  44.9 \small{\textcolor{pos}{(+0.1)}} &
  44.6 \small{\textcolor{neg}{(-0.2)}} &
  45.1 \small{\textcolor{pos}{(+0.3)}} &
  45.0 \small{\textcolor{pos}{(+0.2)}} &
  44.8 \small{($\pm$0.0)} \\
\texttt{gpt-3.5} &
  9.9 &
  \multicolumn{1}{c|}{\textbf{28.5 \small{\textcolor{pos}{(+2.6)}}}} &
  25.9 &
  27.1 \small{\textcolor{pos}{(+1.2)}} &
  \multicolumn{1}{c|}{13.0 \small{\textcolor{neg}{(-12.9)}}} &
  26.5 \small{\textcolor{pos}{(+0.6)}} &
  27.0 \small{\textcolor{pos}{(+1.1)}} &
  \textbf{28.5 \small{\textcolor{pos}{(+2.6)}}} &
  26.5 \small{\textcolor{pos}{(+0.6)}} &
  26.7 \small{\textcolor{pos}{(+0.8)}} &
  27.2 \small{\textcolor{pos}{(+1.3)}} \\
\texttt{gpt-4} &
  9.9 &
  \multicolumn{1}{c|}{46.4 \small{\textcolor{neg}{(-0.5)}}} &
  46.9 &
  46.9 \small{($\pm$0.0)} &
  \multicolumn{1}{c|}{34.6 \small{\textcolor{neg}{(-12.3)}}} &
  47.3 \small{\textcolor{pos}{(+0.4)}} &
  47.5 \small{\textcolor{pos}{(+0.6)}} &
  47.7 \small{\textcolor{pos}{(+0.8)}} &
  47.5 \small{\textcolor{pos}{(+0.6)}} &
  47.8 \small{\textcolor{pos}{(+0.9)}} &
  47.7 \small{\textcolor{pos}{(+0.8)}} \\ \midrule
\textit{\textbf{ARC-Challenge}} &
   &
   &
   &
  \multicolumn{1}{l}{} &
  \multicolumn{1}{l}{} &
   &
   &
   &
   &
   &
   \\ \midrule
\texttt{mistral-7b} &
  24.9 &
  \multicolumn{1}{c|}{53.1 \small{\textcolor{neg}{(-17.9)}}} &
  71.0 &
  70.3 \small{\textcolor{neg}{(-0.7)}} &
  \multicolumn{1}{c|}{47.7 \small{\textcolor{neg}{(-23.3)}}} &
  71.7 \small{\textcolor{pos}{(+0.7)}} &
  71.7 \small{\textcolor{pos}{(+0.7)}} &
  71.6 \small{\textcolor{pos}{(+0.6)}} &
  71.7 \small{\textcolor{pos}{(+0.7)}} &
  71.7 \small{\textcolor{pos}{(+0.7)}} &
  71.6 \small{\textcolor{pos}{(+0.6)}} \\
\texttt{llama-3-8b} &
  25.2 &
  \multicolumn{1}{c|}{44.4 \small{\textcolor{neg}{(-21.8)}}} &
  66.2 &
  52.8 \small{\textcolor{neg}{(-13.4)}} &
  \multicolumn{1}{c|}{41.1 \small{\textcolor{neg}{(-25.1)}}} &
  \textbf{71.3 \small{\textcolor{pos}{(+5.1)}}} &
  \textbf{70.0 \small{\textcolor{pos}{(+3.8)}}} &
  \textbf{70.0 \small{\textcolor{pos}{(+3.8)}}} &
  \textbf{71.6 \small{\textcolor{pos}{(+5.4)}}} &
  \textbf{71.3 \small{\textcolor{pos}{(+5.1)}}} &
  \textbf{71.3 \small{\textcolor{pos}{(+5.1)}}} \\
\texttt{glm-4-9b} &
  24.8 &
  \multicolumn{1}{c|}{69.9 \small{\textcolor{neg}{(-9.7)}}*} &
  79.3 &
  80.1 \small{\textcolor{pos}{(+0.8)}} &
  \multicolumn{1}{c|}{65.1 \small{\textcolor{neg}{(-14.2)}}} &
  \textbf{82.7 \small{\textcolor{pos}{(+3.4)}}} &
  \textbf{82.3 \small{\textcolor{pos}{(+3.0)}}} &
  \textbf{82.0 \small{\textcolor{pos}{(+2.7)}}} &
  \textbf{82.8 \small{\textcolor{pos}{(+3.5)}}} &
  \textbf{83.0 \small{\textcolor{pos}{(+3.7)}}} &
  \textbf{82.7 \small{\textcolor{pos}{(+3.4)}}} \\
\texttt{llama-3-70b} &
  25.3 &
  \multicolumn{1}{c|}{88.9 \small{\textcolor{pos}{(+1.1)}}} &
  87.8 &
  87.9 \small{\textcolor{pos}{(+0.1)}} &
  \multicolumn{1}{c|}{\textbf{80.8 \small{\textcolor{neg}{(-7.0)}}}} &
  88.5 \small{\textcolor{pos}{(+0.7)}} &
  88.4 \small{\textcolor{pos}{(+0.6)}} &
  88.1 \small{\textcolor{pos}{(+0.3)}} &
  88.5 \small{\textcolor{pos}{(+0.7)}} &
  88.4 \small{\textcolor{pos}{(+0.6)}} &
  88.4 \small{\textcolor{pos}{(+0.6)}} \\
\texttt{qwen-2-72b} &
  24.8 &
  \multicolumn{1}{c|}{84.7 \small{\textcolor{neg}{(-1.1)}}} &
  85.8 &
  86.0 \small{\textcolor{pos}{(+0.2)}} &
  \multicolumn{1}{c|}{36.7 \small{\textcolor{neg}{(-49.1)}}} &
  86.3 \small{\textcolor{pos}{(+0.5)}} &
  86.2 \small{\textcolor{pos}{(+1.3)}} &
  85.8 \small{($\pm$0.0)} &
  86.3 \small{\textcolor{pos}{(+0.5)}} &
  86.3 \small{\textcolor{pos}{(+0.5)}} &
  86.2 \small{\textcolor{pos}{(+0.4)}} \\
\texttt{qwen-1.5-110b} &
  24.7 &
  \multicolumn{1}{c|}{87.0 \small{\textcolor{neg}{(-0.7)}}} &
  87.7 &
  88.3 \small{\textcolor{pos}{(+0.6)}} &
  \multicolumn{1}{c|}{53.4 \small{\textcolor{neg}{(-34.3)}}} &
  88.1 \small{\textcolor{pos}{(+0.4)}} &
  88.1 \small{\textcolor{pos}{(+0.4)}} &
  88.0 \small{\textcolor{pos}{(+0.3)}} &
  88.1 \small{\textcolor{pos}{(+0.4)}} &
  88.1 \small{\textcolor{pos}{(+0.4)}} &
  88.1 \small{\textcolor{pos}{(+0.4)}} \\
\texttt{gpt-3.5} &
  25.2 &
  \multicolumn{1}{c|}{78.1 \small{\textcolor{pos}{(+1.2)}}} &
  76.9 &
  77.0 \small{\textcolor{pos}{(+0.1)}} &
  \multicolumn{1}{c|}{29.9 \small{\textcolor{neg}{(-47.0)}}} &
  78.2 \small{\textcolor{pos}{(+1.3)}} &
  77.9 \small{\textcolor{pos}{(+1.0)}} &
  78.2 \small{\textcolor{pos}{(+1.3)}} &
  78.1 \small{\textcolor{pos}{(+1.2)}} &
  77.9 \small{\textcolor{pos}{(+1.0)}} &
  77.9 \small{\textcolor{pos}{(+1.0)}} \\
\texttt{gpt-4} &
  25.0 &
  \multicolumn{1}{c|}{92.9 \small{\textcolor{pos}{(+0.4)}}} &
  92.5 &
  92.8 \small{\textcolor{pos}{(+0.3)}} &
  \multicolumn{1}{c|}{\textbf{87.3 \small{\textcolor{neg}{(-5.2)}}}} &
  92.9 \small{\textcolor{pos}{(+0.4)}} &
  92.7 \small{\textcolor{pos}{(+0.2)}} &
  92.8 \small{\textcolor{pos}{(+0.3)}} &
  92.8 \small{\textcolor{pos}{(+0.3)}} &
  92.8 \small{\textcolor{pos}{(+0.3)}} &
  92.9 \small{\textcolor{pos}{(+0.4)}} \\ \bottomrule
\end{tabular}%
}
\caption{
Overall accuracy results on MMLU, MMLU-Pro and ARC-Challenge benchmarks. `Rand.' and `Dicta.' denote `random' and `dictatorial', respectively. The numbers in parentheses are relative to the \textit{blind dictatorial} baselines. Performance gains are marked in \textcolor{red}{red}, and loss in \textcolor{blue}{blue}. Notable cases are marked in \textbf{bold}. *Results marked with asterisk are calculated utilizing partial profiles (see Appendix~\ref{appendix:stats}).
}
\label{table:main_results}
\end{table*}

\section{A Case Study on MCQA Benchmarks}\label{sec:case study}

\subsection{Experiment Setup}
\paragraph{Datasets} As the primary scope of this study is on decision-making process rather than choice-generation, MCQA benchmarks particularly suits the research interest, for they have alternatives (i.e., choices) predefined.
Following previous studies on benchmarking general performances of LLM agents \cite{10.1145/3526113.3545616, liu2023dynamic, zhang2023exploring, geminiteam2023gemini, jiang2023mistral}, we select MMLU \cite{hendrycks2021measuring}, MMLU-Pro \cite{wang2024mmlupro}, and ARC-Challenge \cite{clark2018think} as the case study testbeds.

\paragraph{Backbone Models}
In an effort to simulate agents built on language models of diverse architectures and parameter sizes, we curate a collection of six open-sourced models, including \texttt{mistral-7b} \cite{jiang2023mistral}, \texttt{glm-4-9b} \cite{zeng2022glm}, \texttt{llama-3-8b/70b} \cite{llama3modelcard} and \texttt{qwen-1.5-72b/110b} \cite{qwen1.5}.
In addition, since most existing LLM-based multi-agent collaboration frameworks employ high performance models with huge parameter sizes to create agents, following their suits, we also test two widely-used proprietary models: \texttt{gpt-3.5} \cite{ouyang2022training} and \texttt{gpt-4} \cite{openai2023GPT4}.
Specifications of selected models can be found in Appendix~\ref{appendix:reproducibility} Table~\ref{table:statistics}.
The temperature of all models are fixed at 0.7 (within a 0.0 to 1.0 range) except for OpenAI models, whose temperatures are maintained at 1.0 (within a 0.0 to 2.0 range).

\paragraph{Metric and Assessment}
For simplicity, we harness unmodified language models as test agents and prefix a short instruction `You are the \{random number\}-th rater' ahead of questions to identify them. A decision ensemble is composed of a designated number of agents built on the same backbone model. 
Every agent is requested to provide a preferential ranking of choices on each question independently.
Having gathered all rankings (i.e., votes) to form a profile, \tool{} outputs collective preferential ranks conforming to selected voting rules.
Uniquely, under \textit{informed} and \textit{mis-informed} dictatorial rules, a `dictator' agent besides the voting ensemble is provided with other agents' votes and then enquired (see \S~\ref{sec:methods}).
As described in \S~\ref{sec:definiation}, given a profile $P$ containing 10 preferential rankings from agents, a voting system $f$ of \tool{} outputs an ordered list $f(P)$ of all choices. 
We consider a question is correctly answered if the first element of the output list match the corresponding gold label.
In accordance with the original setup of MMLU \cite{hendrycks2021measuring}, we implement a 5-shot example prompting that utilises the development sets of the datasets.
All methods take the same preferential ranking format votes except for \textit{range voting} that requires numerical preferential scores in addition to the rankings.

\subsection{Main Results}\label{sec:results}

The 5-run average overall accuracy results are reported in Table~\ref{table:main_results}, and corresponding statistics of valid ranking profiles are detailed in Appendix~\ref{appendix:stats}. 

\paragraph{Random Baselines and Range Voting} 
The accuracies for random baselines hover around 25.0 for the 4-choice MMLU and ARC-Challenge, and approximately 10.0 for the 10-choice MMLU-Pro. These figures confirm a balanced distribution of correct choices within the test sets. Most models, especially the smaller ones, exhibit inferior performance when implementing score-based \textit{range voting} (i.e., cardinal ranking) compared to ordinal ranking methods. However, \texttt{llama-3-70b}, \texttt{gpt-3.5}, and \texttt{gpt-4} are exceptions, as their \textit{range voting} outcomes exceed those of \textit{blind dictatorial}.

\paragraph{Dictatorial Methods}
The colored numbers in Table~\ref{table:main_results} indicate results relative to \textit{blind dictatorial}, which serves as the baseline for comparing. Although most models perform better under \textit{informed dictatorial} than under \textit{blind dictatorial}, they do not outperform other ordinal ranking methods.

It should be noted that \textit{informed dictatorial} cost more than voting-based methods computationally, since it necessitates a complete ballot profile from the ensemble, in addition to the `dictator'. The subpar performance of \textit{informed dictatorial} implies that a `dictator' agent is unable to utilize the information from ensemble ballots more effectively than the voting systems.

As anticipated, the significantly reduced accuracies under \textit{mis-informed dictatorial} demonstrate the detrimental effect of providing the `dictator' with random ballots. Remarkably, \texttt{glm-4-9b} and \texttt{gpt-4} exhibit a relatively minor decline compared to other models across the three datasets, indicating their resilience to misleading information.

\paragraph{Ordinal Ranking Methods}
It is consistently observed that the application of voting-based ordinal ranking methods, even those as straightforward as \textit{plurality}, results in accuracies that match or surpass those achieved by \textit{blind dictatorial}. The extent of improvement varies depending on the specific model in question. Notably, models built on smaller models (<10B) and those within the GPT series exhibit substantial performance enhancements when electoral CDM methods are employed, in stark contrast to medium models (10-110B).

For MMLU benchmark, the adoption of a voting method leads to average accuracy increases of approximately 2.9\%, 4.9\%, and 6.5\% for \texttt{glm-4-9b}, \texttt{gpt-3.5}, and \texttt{gpt-4}, respectively. Given that MMLU-Pro is a 10-choice test, the relative improvements due to CDM may appear less pronounced. Nonetheless, \texttt{llama-3-8b} and \texttt{glm-4-9b} still register noticeable accuracy gains under voting methods. In particular, \textit{minimax} and \textit{ranked pairs} methods demonstrate robustness, showing positive effects on all models across the three benchmarks.

These findings call for a reassessment on existing LLM-agent collaboration frameworks, particularly regarding \textit{the extent to which the impacts of their proposed systems may be attributed to the implementation of specific CDM methods.} However, it is also observed that some CDM methods exhibit marginal and indistinct differences in performance on certain models, warranting further detailed examination.

\subsection{Analysis and Discussion}\label{sec:analysis}
Having observing that the agents build on \texttt{gpt-3.5} and \texttt{gpt-4} demonstrate the most significant improvement under ordinal ranking methods, we follow up with additional inquires and analyses.

\paragraph{Minimum Effective Voting Quorum}
Firstly, we pose an intuitive question regarding the voting quorum: \textit{What is the minimum number of agents to compose an effective decision group?}

\begin{figure}[h]
\centering
\includegraphics[width=\columnwidth]{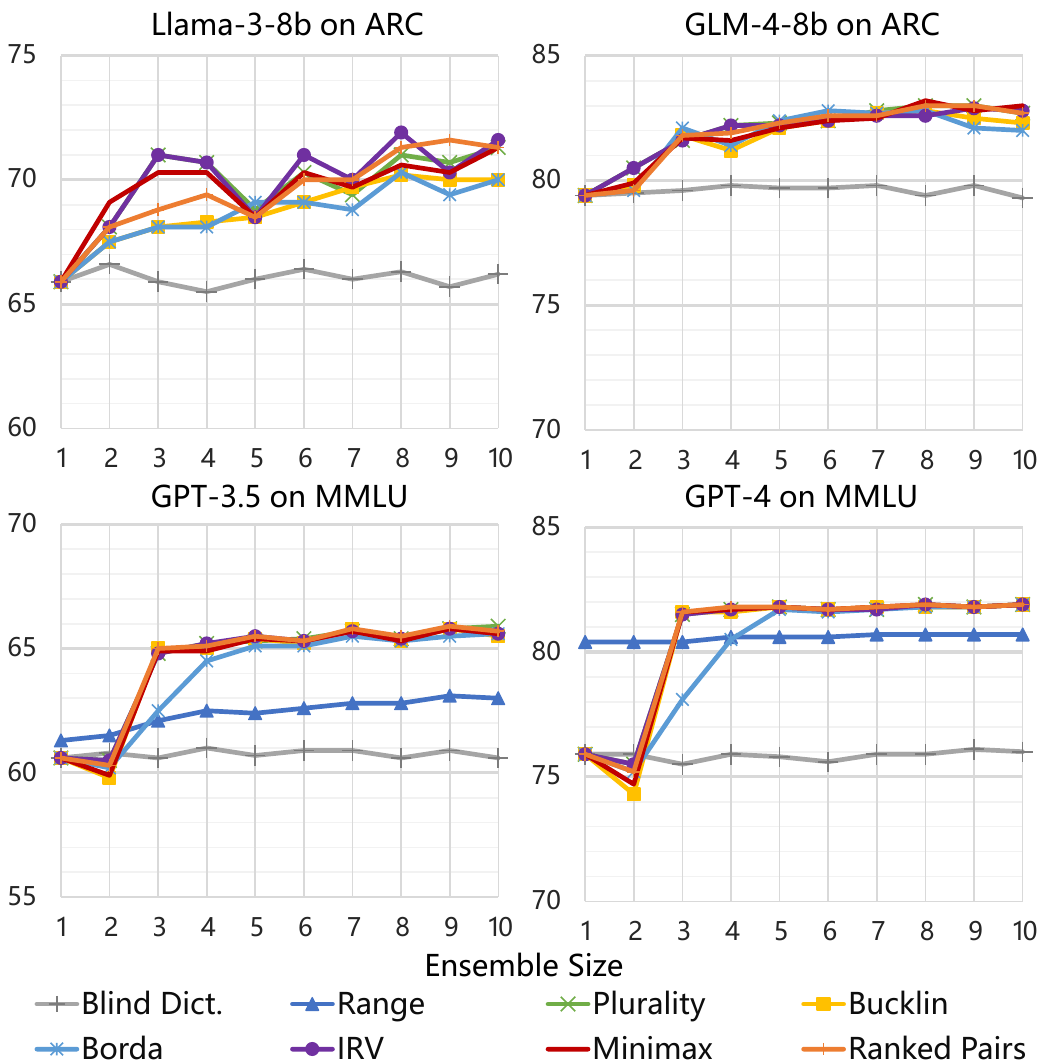}
\caption{
Accuracy comparison of voting ensembles of different sizes built on the same backbone models. The \textit{Range} results of \texttt{glm-4-9b} is excluded for insufficient profiles (see Appendix~\ref{appendix:stats}).
}
\label{fig:ensemble}
\end{figure}

Figure~\ref{fig:ensemble} illustrates the notable impact on accuracy when varying the number of voting agents, using \texttt{llama-3-8b} and \texttt{glm-4-9b} on the ARC-Challenge dataset, and \texttt{gpt-3.5} and \texttt{gpt-4} on the MMLU benchmark. Overall, most CDM methods start exhibiting significant improvements and surpass the \textit{blind dictatorial} baseline in situations involving more than two agents, where a majority can be established.

For GPT models, noticeable drops occurs when the voting group increases to two.
\textit{Borda} takes a few more agents to reach the plateau, which is likely attributed to its ballot weight scale that is based on the number of choices (4 in our case).
\textit{Range voting} starts higher yet stabilizes lower than other methods.
Surprisingly, for \texttt{gpt-4}, simply requiring a range vote rather than ordinal preferential vote greatly increases its judgement even without multiple agents! 
However, the results of \textit{range voting} vary slightly when increasing number of voting agents, demonstrating a ensemble-size-independent property that is not seen on \texttt{gpt-3.5}.
In particular, \texttt{llama-3-8b} shows the most variance when applying different CDM, mostly due to a smaller number of valid profiles (see Appendix~\ref{appendix:stats}).
Nonetheless, since the ensemble size directly impacts the required computational resources, a consideration of cost-benefit trade-offs is essential.

\paragraph{Robustness against Unreliable Agents}
The voting quorum scenario presupposes that all agents can accurately express their preferences. However, one might wonder: \textit{What if LLM agents are unreliable (i.e., malfunctioning or incapable)?} An extra advantage of involving more agents in decision-making is the increased robustness against a single point of failure. To assess the resilience of various methods to unreliable voters, we incrementally replaced the voting ensemble of 10 fully functional agents with unreliable ones who cast random votes.

\begin{figure}[h]
\centering
\includegraphics[width=\columnwidth]{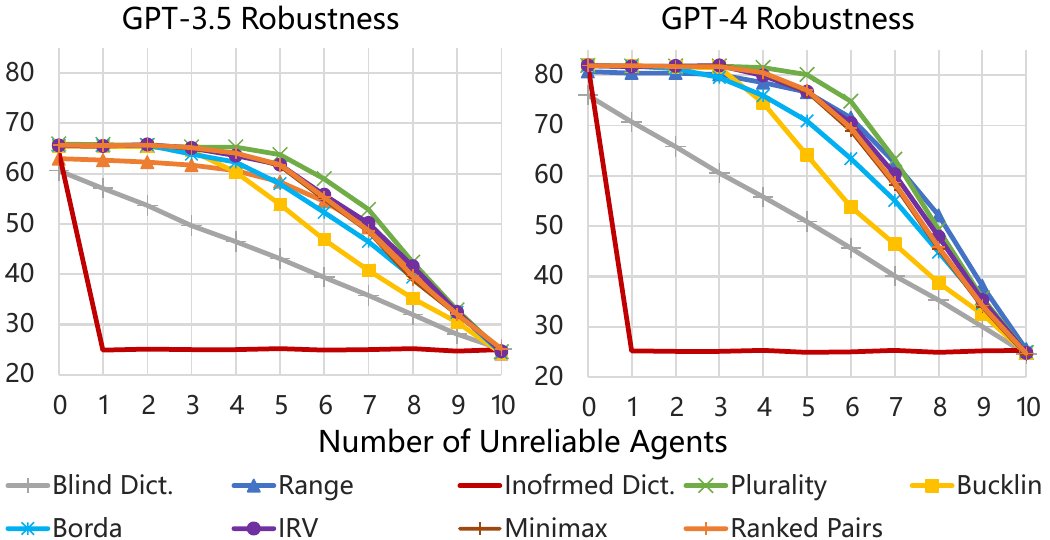}
\caption{
Accuracy impact of increasing number of unreliable agents built on \texttt{gpt-3.5} and \texttt{gpt-4}. 
}
\label{fig:robust}
\end{figure}

Figure~\ref{fig:robust} depicts the performance of compromised voting ensembles under different voting rules.
Most voting methods maintain their integrity until the number of unreliable agents reaches 4, and then their accuracies converges to the random baseline at 25\%.
As anticipated, \textit{informed dictatorial} is the first to collapse, since the entire system fails once the `dictator' is incapable of making a reasonable judgment (\textit{utilitarian} methods relying on single external utility-calculation module would be the same case).
Contrary to expectations, \textit{plurality} exhibits a commendable robustness compared to more sophisticated methods.

\begin{figure}[h]
\centering
\includegraphics[width=\columnwidth]{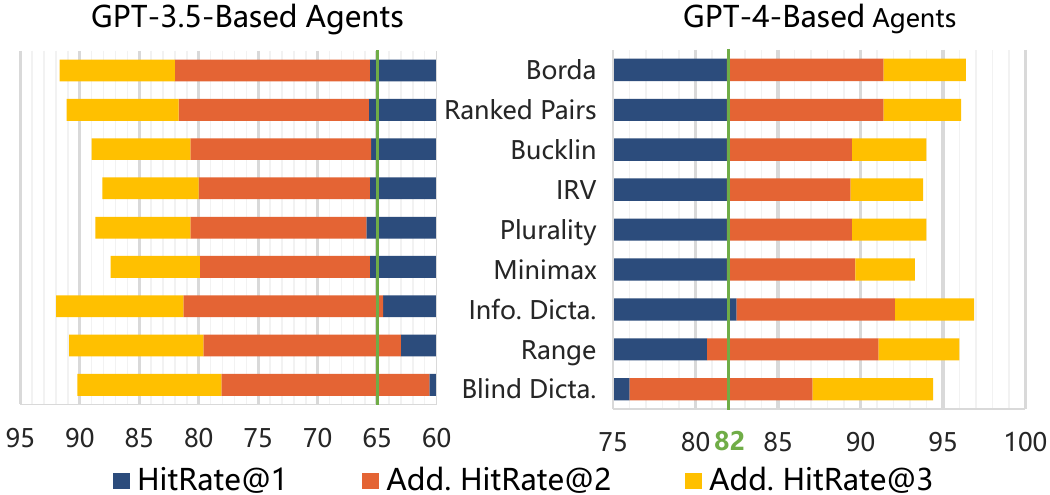}
\caption{
Hit-rate@$k$ comparison of different voting rules utilising ballots given by voting agents. \textcolor{green}{Green} lines are drawn to highlight similar hit-rate@1. 
}
\label{fig:hitrate}
\end{figure}

\paragraph{Difference in Hit-Rate@$K$}
Let hit-rate@$k$ denotes a cumulative accuracy of taking the first $k$ preferences of an answer. We find that although a few methods yield seemingly even performance gains, they are distinguishable in terms of hit-rate@$k$, as illustrated in Figure~\ref{fig:hitrate}.
Notably, despite being robust against unreliable agent, \textit{plurality} falls short in scenarios where the elimination of the worst choices is of the higher priority than the selection of the best.
On the other hand, \textit{Borda}, \textit{ranked pairs}, and \textit{informed dictatorial} methods have the strongest discriminant power on excluding the wrong choices.
Intriguingly, while \textit{blind dictatorial} performs poorly on the first choice, its hit-rate@3 surpasses some electoral methods.

\begin{figure}[h]
\centering
\includegraphics[width=\columnwidth]{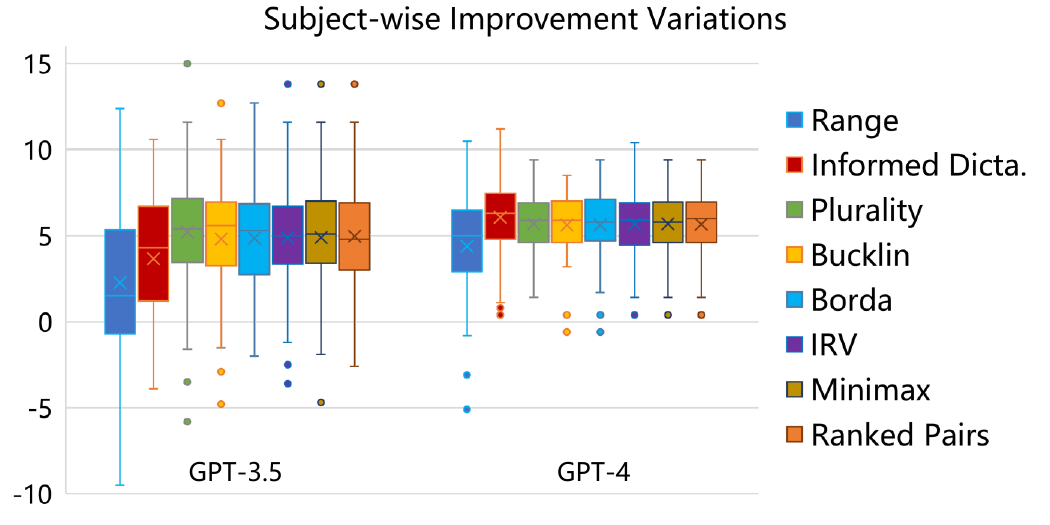}
\caption{Box plots of subject-wise accuracy improvement variations under different CDM methods.
}
\label{fig:subjectwise_box}
\end{figure}

\paragraph{Subject-wise Performance Improvements}
Inspecting the subject-wise results in Figure~\ref{fig:subjectwise_box}, we find that, under the same voting method, the performance gains are not evenly distributed across disciplines.
Taking \textit{plurality} for instance, the subject-wise accuracy improvements range from -5.8\% to +15.0\% for \texttt{gpt-3.5} and from 1.4\% and 9.4\% for \texttt{gpt-4}. 

\begin{figure}[h]
\centering
\includegraphics[width=\columnwidth]{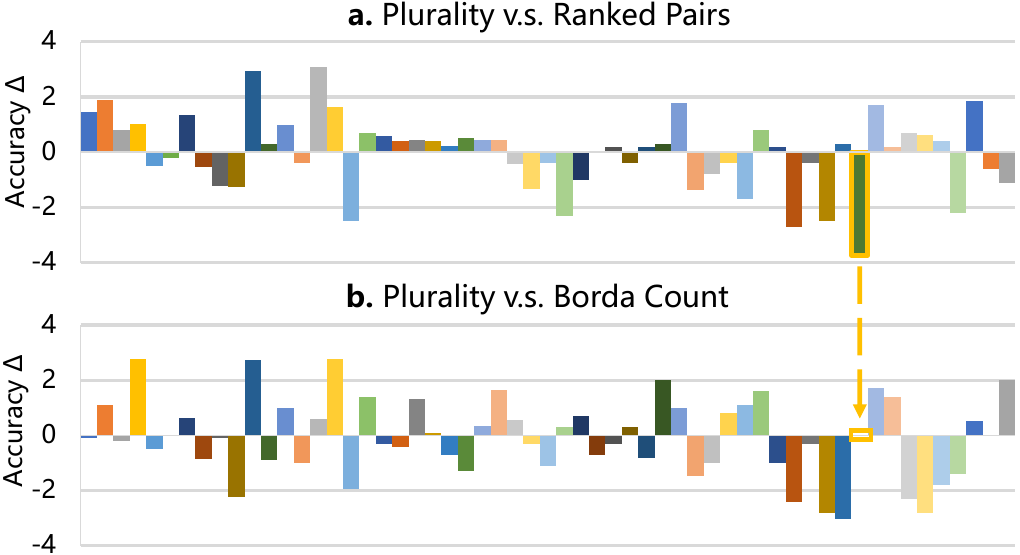}
\caption{
An example of CDM impacts on subject-wise accuracies when holding the model fixed (\texttt{gpt-4} in this case). Each bar denotes a subject-wise accuracy difference between the compared CDM method pair. 
}
\label{fig:subject_example}
\end{figure}

Vice versa, for the same discipline, the impacts under different CDM methods vary as well.
For instance, the dark green bar outlined by golden border in Figure~\ref{fig:subject_example}(a) indicates that \textit{ranked pairs} is -3.7\% less accurate than \textit{plurality} on `professional accounting'. 
Conversely, the corresponding one in Figure~\ref{fig:subject_example}(b) shows no difference between \textit{plurality} and \textit{Borda Count} on the same subject.

Above observations again support our motivation for diversifying decision-making methods in LLM-based multi-agent collaboration.

\section{Conclusion and Future Work}\label{sec:conclusion}
In the midst of the expanding research on LLM-based agents, we have surveyed \surveyed{} multi-agent collaboration frameworks and revealed a significant lack of diversity in CDM.
We have scrutinized popular CDM methods and indicate their fundamental limitations through a social choice theory perspective.
Aiming to diversify the current CDM landscape, we have drew inspiration from human societal practices and explored various CDM methods in an empirical case study across multiple benchmarks.
Our experiments have produced a wealth of observations and insights, demonstrating how such diversification can illuminate the study of collective behaviors in LLMs.

Our study also opens up numerous avenues for future research.
For instance, matching specific tasks with appropriate CDM methods to enhance agent decision-making quality holds promising practical value. 
Moreover, since social choice theory addresses collective preferences, we expect that it could inspire broader interdisciplinary NLP research, particularly language model alignment and aggregation.
\newpage

\section*{Limitations}\label{sec:limitations}
\paragraph{Multi-Choice Question-Answering (MCQA) Benchmarks as CDM Testbed}
while the experiments on MMLU, MMLU-Pro and ARC yield notable and insightful observations, we acknowledge that MCQA is not fully aligned with collective decision-making. Foremost, LLM have demonstrated inconsistency in multi-choice ranking task \cite{zhao-etal-2024-measuring}.
Secondly, most MCQA benchmarks have predetermined `correct' answers; however, CDM processes can also be relevant in scenarios where there is no absolute right or wrong. For instance, measuring bias in LLM agents involves aggregating the `preferences' of individual agents, where no objectively `correct' choices exist. Therefore, an additional avenue for future work could involve constructing a benchmark that measures preference representativeness rather than one based on true-or-false judgments. 

\paragraph{Self-contained Testing}
All experiments are self-contained systems of sole backbone model. In other words, we do not test any ensemble containing voting agents built on different LLMs, which could be another future direction.

\paragraph{Unexhausted Inclusion of Voting Strategies in \tool{}}
Although we attempted to cover common modern electoral systems, the CDM method list of \tool{} is not exhaustive.
For instance, in an effort to keep the module compact and lightweight, we do not include compound mechanisms that combine multiple voting strategies. However, such mechanisms are achievable by arranging a pipeline of multiple \tool{} modules if so wish.

\paragraph{`Voting Tax'}
The `voting tax' of electoral CDM methods refers to the computation cost of implementing  such methods. The tax is composed of two parts: agent actions and ballot processing.
Agent actions takes largest proportion as operating LLM agents is highly costly.
The cost of inter-agent communication should be taken into consideration as well. Compared with vast computational resources required by model inference, ballot processing part consumes much minor.

Another aspect to consider the cost-benefit trade-offs is the `participation' in decision-making.
Human voters could feel certain degree of fulfillment by participation alone regardless of results, as they have expressed their preferences in social decision-making processes. LLM agents, however, can not benefit through participation. This distinction makes voting population factor in LLM-agent CDM a totally utilitarian one.

\section*{Broader Impacts and Ethical Considerations}
The purpose of this research is to explore the possibilities of implementing diverse collective decision-making methods among LLM-based agents. 
However, this study does not support nor encourage any attempt to utilize LLM agents as representatives to replace human judgment in real-world democratic decision-making processes. 

\section*{Acknowledgements}
We are grateful to the anonymous reviewers and editors for their constructive feedback. Special thanks go to our colleague Jack for his support in deploying and managing the inference services for the open-source models used in this study.

\bibliography{acl_latex}

\appendix
\addcontentsline{toc}{section}{Appendices}
\renewcommand{\thesubsection}{\Alph{subsection}}

\section{Reproducibility Statement}\label{appendix:reproducibility}
We employ 8 backbone models for the experiments.
\texttt{gpt-3.5} and \texttt{gpt-4} are commercially available proprietary models.
Specifically, we adopt the snapshot models \texttt{gpt-3.5-turbo-1106} and \texttt{gpt-4-0125-preview}. 
As for the open-source models, we adopt \texttt{Mistral-7B-v0.3}, \texttt{glm-4-9b-chat}, \texttt{Llama-3-8B/70B-Instruct}, and \texttt{Qwen1.5-72B/110B-Instruct}. The sources of above models are listed in Table~\ref{table:models}.

\begin{table}[h]
\centering
\resizebox{\columnwidth}{!}{%
\begin{tabular}{@{}ll@{}}
\toprule
\textbf{Models}        & \textbf{Sources}                   \\ \midrule
\texttt{mistral-7b}   & \url{https://huggingface.co/mistralai/Mistral-7B-Instruct-v0.3}   \\
\texttt{llama-3-8b}   & \url{https://huggingface.co/meta-llama/Meta-Llama-3-8B-Instruct}  \\
\texttt{glm-4-9b}     & \url{https://huggingface.co/THUDM/glm-4-9b-chat}                  \\
\texttt{llama-3-70b}  & \url{https://huggingface.co/meta-llama/Meta-Llama-3-70B-Instruct} \\
\texttt{qwen1.5-72b}    & \url{https://huggingface.co/Qwen/Qwen1.5-72B-Chat}       \\
\texttt{qwen1.5-110b} & \url{https://huggingface.co/Qwen/Qwen1.5-110B-Chat}               \\
\texttt{gpt-3.5-turbo} & \url{https://platform.openai.com/} \\
\texttt{gpt-4}         & \url{https://platform.openai.com/} \\ \bottomrule
\end{tabular}%
}
\caption{
Specification and sources of evaluated models.
}
\label{table:models}
\end{table}

\section{Surveyed LLM-based Multi-Agent Collaboration Frameworks and Systems}\label{appendix:mas}
\begin{table}[H]
\resizebox{\columnwidth}{!}{%
\begin{tabular}{@{}lll@{}}
\toprule
CDM Method  & Systems and Frameworks             & Note                \\ \midrule
\multirow{23}{*}{Dictatorial} &
  \citet{xiong-etal-2023-examining} &
  Assigned role \\
            & \citet{wu2023autogen}              & Assigned role       \\
            & \citet{hao2023chatllm}             & Assigned role       \\
            & \citet{liu2023dynamic}             & Assigned role       \\
            & \citet{li2023camel}                & Assigned role       \\
            & \citet{zhang2023proagent}          & Assigned role       \\
            & \citet{nair2023dera}               & Assigned role       \\
            & \citet{talebirad2023multiagent}    & Assigned role       \\
            & \citet{liang2023encouraging}       & Assigned role       \\
            & \citet{tang2023causalgpt}          & Assigned role       \\
            & \citet{qian2023communicative}      & Assigned role       \\
            & \citet{sun2023corex}               & Assigned role       \\
            & \citet{chen2023gamegpt}            & Assigned role       \\
            & \citet{jinxin2023cgmi}             & Assigned role       \\
            & \citet{li2023traineragent}         & Assigned role       \\
            & \citet{fang2024multiagent}         & Assigned role       \\
            & \citet{tang2024collaborative}      & Assigned role       \\
            & \citet{hang2024cca}                & Assigned role       \\
            & \citet{darcy2024marg}              & Assigned role       \\
            & \citet{hua2024trustagent}          & Assigned role       \\
            & \citet{wang2024xuatcopilot}        & Assigned role       \\
            & \citet{li2023metaagents}           & Assigned role       \\
            & \citet{chen2023autoagents}         & Oligarchy           \\ \midrule
\multirow{15}{*}{\begin{tabular}[c]{@{}l@{}}No CDM \\or\\ Unspecified \end{tabular}} &
  \citet{he-etal-2023-lego} &
  Decentralized team \\
            & \citet{li-etal-2023-theory}        & Decentralized team  \\
            & \citet{nakajima2023babyagi}        & Decentralized team  \\
            & \citet{NI2024102131}               & Human judgement     \\
            & \citet{ghafarollahi2024protagents} & Human judgement     \\
            & \citet{wang2023macsql}             & Linear workflow     \\
            & \citet{ding2023designgpt}          & Linear workflow     \\
            & \citet{hong2023metagpt}            & Linear workflow     \\
            & \citet{rasheed2024codepori}        & Linear workflow     \\
            & \citet{wei2024editable}            & Linear workflow     \\
            & \citet{liu2023training}            & Scenario simulation \\
            & \citet{park2023generative}         & Scenario simulation \\
 &
  \citet{ghaffarzadegan2023generative} &
  Scenario simulation \\
            & \citet{hua2023war}                 & Scenario simulation \\
            & \citet{zhang2024speechagents}      & Scenario simulation \\ \midrule
\multirow{13}{*}{Plurality} &
  \citet{du2023improving} &
  Consensus \\
            & \citet{wang2023unleashing}         & Consensus           \\
            & \citet{chen2023agentverse}         & Consensus           \\
            & \citet{chen2023multiagent}         & Consensus           \\
            & \citet{li2023tradinggpt}           & Consensus           \\
            & \citet{shi2023cooperation}         & Game rule           \\
            & \citet{stepputtis-etal-2023-long}  & Game rule           \\
            & \citet{xu2023exploring}            & Game rule           \\
            & \citet{chan2023chateval}           & Relative majority   \\
            & \citet{xu2023reasoning}            & Relative majority   \\
            & \citet{zhang2023exploring}         & Relative majority   \\
            & \citet{li2024agents}               & Relative majority   \\
            & \citet{hamilton2023blind}          & Scenario simulation \\ \midrule
Utilitarian & \citet{jarrett2023language}        &                     \\ \bottomrule
\end{tabular}%
}
\caption{
Full list of \surveyed{} surveyed LLM-based multi-agent collaboration works.
}
\label{table:mas}
\end{table}

\section{Main Experiment Statistics}\label{appendix:stats}
For MMLU and MMLU-Pro datasets, we curate subject-wise balanced test subsets by selecting first 100 cases of each subject (i.e., discipline). Thus, the subset contains 5,700 questions for MMLU and 1,400 for MMLU-Pro. Regarding ARC-Challenge, the whole test set of 1,172 cases are used.

We consider a profile to be valid if (1) the profile comprises ballots from all voting agents, and (2) every ballot includes a complete and non-duplicated ranked list of choices and matches the instructed format. Only valid profiles are forwarded to GEDI and processed. The statistics of main experiments are summarized in Table~\ref{table:statistics}.

\begin{table}[h]
\resizebox{\columnwidth}{!}{%
\begin{tabular}{@{}lcccc@{}}
\toprule
\textbf{MMLU} &
  Range &
  \begin{tabular}[c]{@{}c@{}}Ordinal\\ Ranking\end{tabular} &
  Informed &
  \begin{tabular}[c]{@{}c@{}}Mis\\ -informed\end{tabular} \\ \midrule
\texttt{mistral-7b}    & 2379      & 4788 & 5422 & 5596 \\
\texttt{llama-3-8b}    & 1253      & 1946 & 4961 & 5121 \\
\texttt{glm-4-9b}      & 332       & 3470 & 5502 & 5447 \\
\texttt{llama-3-70b}   & 3909      & 5110 & 5576 & 5435 \\
\texttt{qwen1.5-72b}   & 4642      & 5657 & 5698 & 5700 \\
\texttt{qwen1.5-110b}  & 5569      & 5625 & 5685 & 5692 \\
\texttt{gpt-3.5-trubo} & 5627      & 5397 & 5569 & 5679 \\
\texttt{gpt-4}         & 5515      & 5572 & 5539 & 5648 \\ \midrule
\textbf{MMLU-Pro} &
  \multicolumn{1}{l}{} &
  \multicolumn{1}{l}{} &
  \multicolumn{1}{l}{} &
  \multicolumn{1}{l}{} \\ \midrule
\texttt{mistral-7b}    & 554       & 564  & 1180 & 1382 \\
\texttt{llama-3-8b}    & 3 (1161*) & 261  & 1162 & 1255 \\
\texttt{glm-4-9b}      & 3 (1359*) & 376  & 1294 & 1323 \\
\texttt{llama-3-70b}   & 1239      & 1293 & 1396 & 1394 \\
\texttt{qwen1.5-72b}   & 388       & 831  & 1284 & 1383 \\
\texttt{qwen1.5-110b}  & 632       & 1138 & 1319 & 1399 \\
\texttt{gpt-3.5-turbo} & 655       & 1283 & 1400 & 1400 \\
\texttt{gpt-4}         & 1375      & 1386 & 1399 & 1397 \\ \midrule
\textbf{ARC-Challenge} &           &      &      &      \\ \midrule
\texttt{mistral-7b}    & 373       & 1033 & 1131 & 1163 \\
\texttt{llama-3-8b}    & 252       & 317  & 1024 & 1043 \\
\texttt{glm-4-9b}      & 1 (1096*) & 1081 & 1153 & 1159 \\
\texttt{llama-3-70b}   & 901       & 1135 & 1172 & 1172 \\
\texttt{qwen1.5-72b}   & 1068      & 1172 & 1172 & 1172 \\
\texttt{qwen1.5-110b}  & 1166      & 1169 & 1171 & 1171 \\
\texttt{gpt-3.5-trubo} & 1172      & 1172 & 1172 & 1172 \\
\texttt{gpt-4}         & 1172      & 1172 & 1171 & 1172 \\ \bottomrule
\end{tabular}%
}
\caption{
Overview statistics of output profile validity of different models on tested datasets. Specifically, since voting profiles of all non-dictator agents is a prerequisite for \textit{informed dictatorial}, we filter out incomplete profiles of other agents before feeding them to the `dictator'. Therefore, the valid profile counts for \textit{informed dictatorial} are bound to be fewer than the original ones. *Since \texttt{llama-3-8b} and \texttt{glm-4-9b} yield too few complete profiles under \textit{range voting} for certain benchmarks, we utilize incomplete profiles with valid ballots to calculate those accuraies in the main experiments.
}
\label{table:statistics}
\end{table}

\newpage

\section{Several CDM Method Criteria Examples}\label{appendix:criteria_example}

\begin{figure}[H]
\centering
\includegraphics[width=0.85\columnwidth]{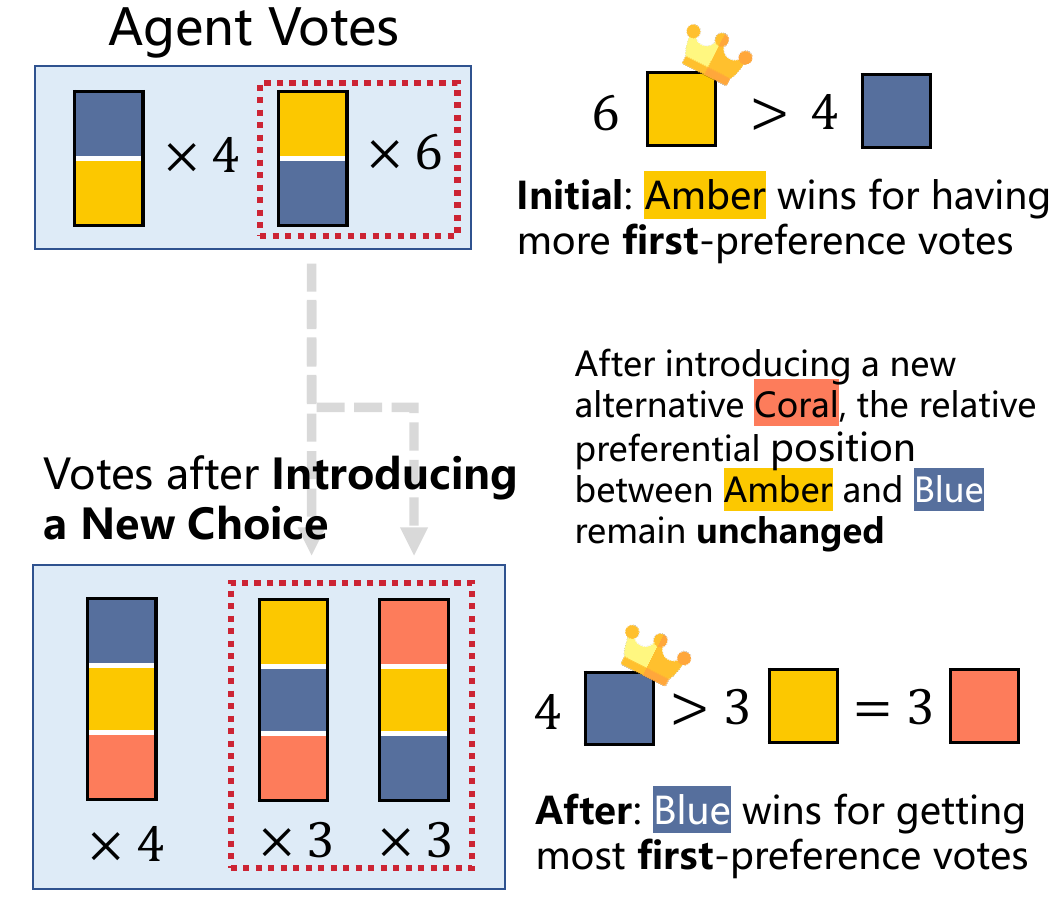}
\caption{
An example of \textit{plurality voting} (the choice with the most first-preference votes wins) violating \textit{Independence from Irrelevant Alternatives (IIA)} criterion. 
Initially, \colorbox{amber}{Amber} wins for two more first-preference votes. However, after introducing a new choice \colorbox{coral}{Coral}, while the relative preferential position between \colorbox{amber}{Amber} and \colorbox{navy}{\textcolor{white}{Blue}} remain unchanged, \colorbox{navy}{\textcolor{white}{Blue}} wins for getting one more first-preference vote than other two options.
}
\label{fig:plurality_IIA}
\end{figure}

\begin{figure}[H]
\centering
\includegraphics[width=0.85\columnwidth]{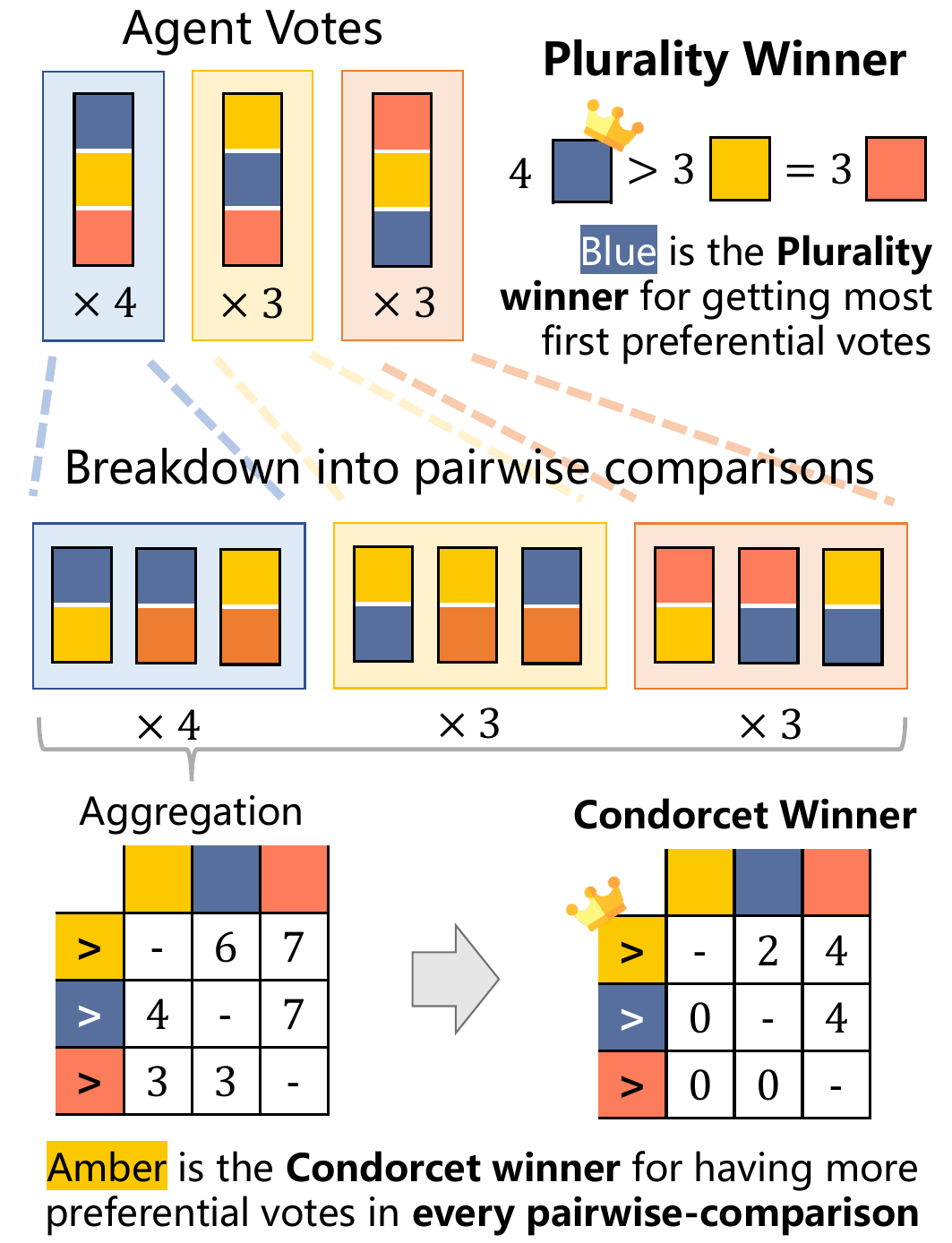}
\caption{
An example of \textit{plurality voting} violating \textit{Condorcet} criterion.
While \colorbox{navy}{\textcolor{white}{Blue}} is the plurality winner for getting the most first-preference votes, \colorbox{amber}{Amber} is actually the \textit{Condorcet Winner}, meaning that \colorbox{amber}{Amber} gets more preferential votes in every pairwise-comparison with other alternatives. This misalignment is due to that \textit{plurality voting} takes only first-preference into account.
}
\label{fig:plurality_condorcet}
\end{figure}

\begin{figure}[h]
\centering
\includegraphics[width=0.85\columnwidth]{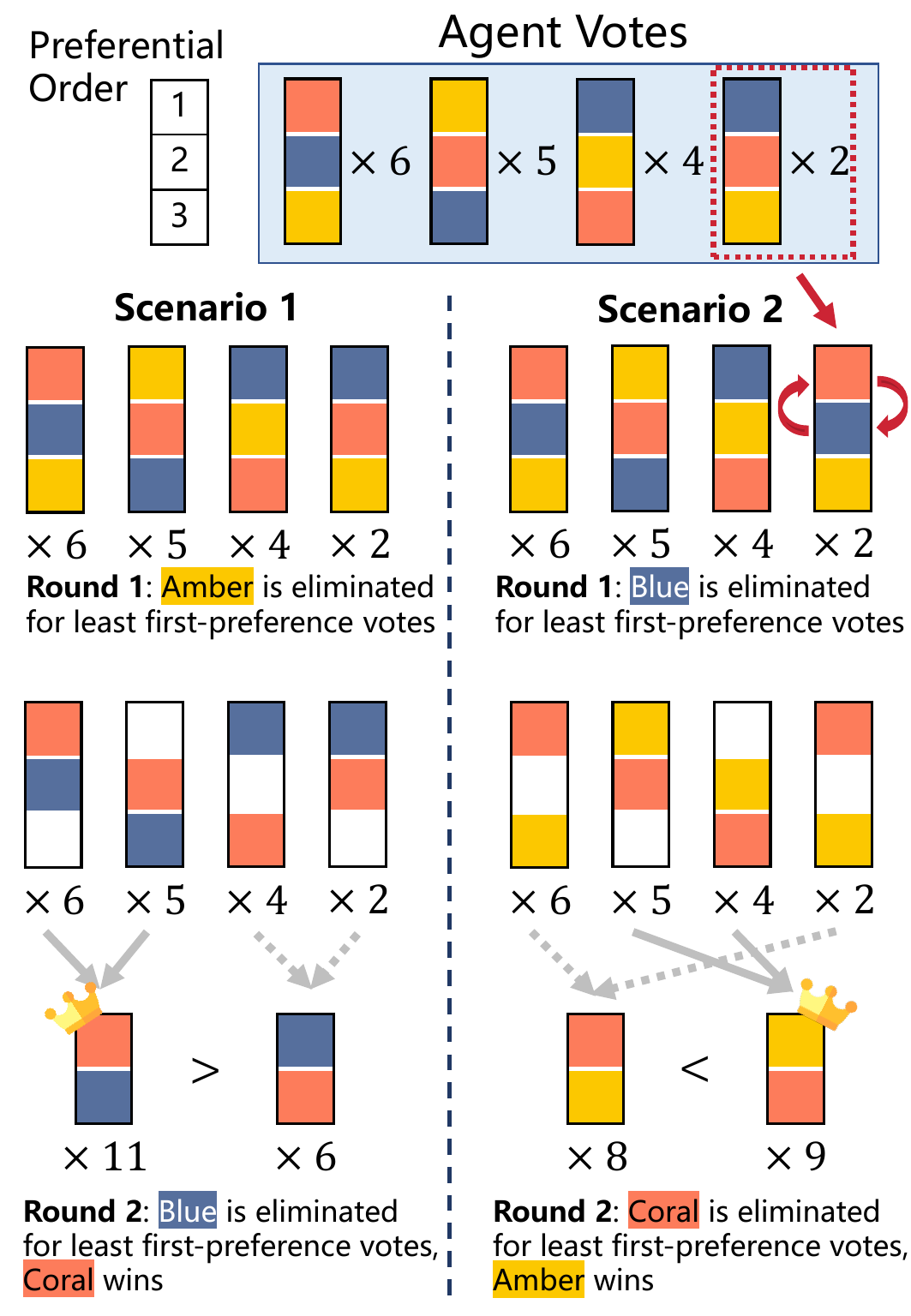}
\caption{
An example of violating \textit{monotonicity} criterion \cite{woodall1997monotonicity} in preferential \textit{instant-runoff voting (IRV)}: repeatedly eliminating the option with the least first preference votes each round until a winner is left. 
In Scenario 2 (right), two agents alter their votes by putting \colorbox{coral}{Coral} first, but this `favorable' action actually \textbf{harms} \colorbox{coral}{Coral} and prevent it from supposed winning.
}
\label{fig:irv_monotonic}
\end{figure}

\begin{figure}[h]
\centering
\includegraphics[width=0.85\columnwidth]{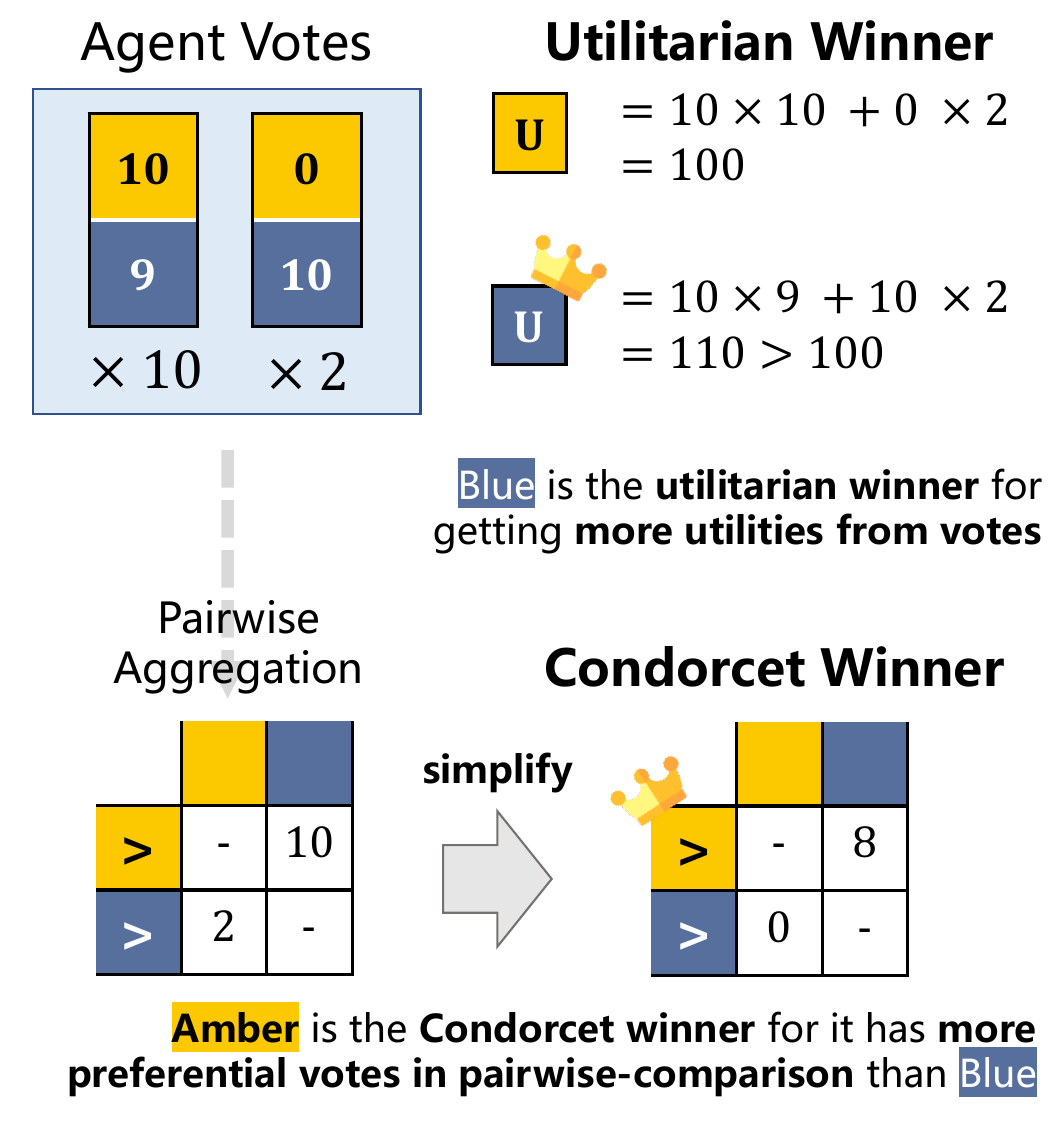}
\caption{
An example of \textit{utilitarian} method violating \textit{Majority} and \textit{Condorcet} criteria.
\colorbox{navy}{\textcolor{white}{Blue}} is the utilitarian winner for getting more utilities from votes, but \colorbox{amber}{Amber} is preferred by the majority of the agents (10 out of 12). In addition, \colorbox{amber}{Amber} is also the \textit{Condorcet Winner}, meaning that \colorbox{amber}{Amber} gets more preferential votes in pairwise-comparison with other alternatives.
}
\label{fig:utilitarian_majority}
\end{figure}

\end{document}